%% file: main.tex
\documentclass[11pt]{article}

\usepackage[preprint]{acl}

\usepackage{times}
\usepackage{latexsym}

\usepackage[T1]{fontenc}

\usepackage[utf8]{inputenc}

\usepackage{microtype}

\usepackage{inconsolata}

\usepackage{graphicx}
\usepackage{xcolor}
\definecolor{AlgGreen}{RGB}{34,139,34}

\newcommand{\algcmt}[1]{\hfill \textcolor{AlgGreen}{\ensuremath{\triangleright}\,#1}}
\usepackage{amsmath}
\usepackage{mathtools}
\usepackage{amssymb}
\usepackage{algorithm}
\usepackage{algorithmic}
\usepackage{bbm}
\usepackage{booktabs}

\usepackage[most]{tcolorbox}
\usepackage{multirow}
\usepackage{subcaption}

\newtcolorbox{promptbox}[1]{
  colback=gray!10,
  colframe=gray!70!black,
  title=#1,
  fonttitle=\bfseries,
  arc=3mm,
  boxrule=0.8pt,
  left=6pt,
  right=6pt,
  top=6pt,
  bottom=6pt
}

\newtcolorbox{answerbox}[1]{
  colback=blue!5,
  colframe=blue!80!black,
  title=#1,
  fonttitle=\bfseries,
  arc=3mm,
  boxrule=1pt,
  left=6pt,
  right=6pt,
  top=6pt,
  bottom=6pt
}

\newtcolorbox{examplebox}[2][]{
  colback=gray!4,
  colframe=gray!50!black,
  title={#2},
  fonttitle=\bfseries,
  arc=2mm,
  boxrule=0.6pt,
  left=6pt,
  right=6pt,
  top=6pt,
  bottom=6pt,
  breakable,
  #1
}

\title{Gumbel Machine: Counterfactual Student Writing Generation\\via Gumbel Noise Steering}

\author{
\bf
Hunter McNichols$^{1}$,
Alexander Scarlatos$^{1}$,
Mihai Dascalu$^{2}$,
Danielle McNamara$^{3}$,
Andrew Lan$^{1}$ \\
$^{1}$University of Massachusetts Amherst $^{2}$ University Politehnica of Bucharest 
$^{3}$Arizona State University \\
\texttt{wmcnichols@umass.edu, ajscarlatos@umass.edu} \\
\texttt{mihai.dascalu@upb.ro,} 
\texttt{Danielle.McNamara@asu.edu,}
\texttt{andrewlan@cs.umass.edu}
}

\begin{document}
\maketitle
\begin{abstract}
An effective method of teaching across disciplines is to provide examples of high-quality work.
However, an example may be significantly different from a student's current work, making it challenging for them to emulate. An ideal learning demonstration is a \textit{counterfactual} version of the student work, an improved version that is still similar to their own.
Existing automated approaches for counterfactual text generation using Large Language Models (LLMs) result in domain-specific systems that are difficult to translate into practical applications.
We present the Gumbel Machine, a flexible, modular approach to generating counterfactuals that leverages LLM instruction-following capabilities while encouraging similarity to a reference factual text.  
Central to our approach is a novel, controlled decoding algorithm, $\beta$-Hindsight control, which uses latent randomness as a tunable similarity control mechanism during counterfactual generation.
Experiments on datasets of student writing, scored on various criteria, demonstrate the effectiveness of our approach at generating counterfactuals both rubric-consistent and similar to a reference.


\end{abstract}

\section{Introduction}

Providing feedback on student work, especially open-ended work, is an essential aspect of education since it helps learners identify specific areas for improvement \cite{hattie_power_2007}. There are two common ways to provide feedback: using a set of rubrics to score student work and highlight areas for improvement, and giving an example of model work. The latter faces a problem: while examples show students what exceptional work looks like, the model example might differ considerably from the student's work in terms of style, approach, or personal tendencies. This potential mismatch can make it difficult for students to see how the example can be applied to their work, thereby limiting the effectiveness of such feedback. Ideally, when receiving feedback, a student would receive a personalized, \emph{counterfactual} version of their work, which indicates small but pedagogically informative changes they can make.


\begin{figure}[!t]
    \centering
    \includegraphics[width=1\linewidth]{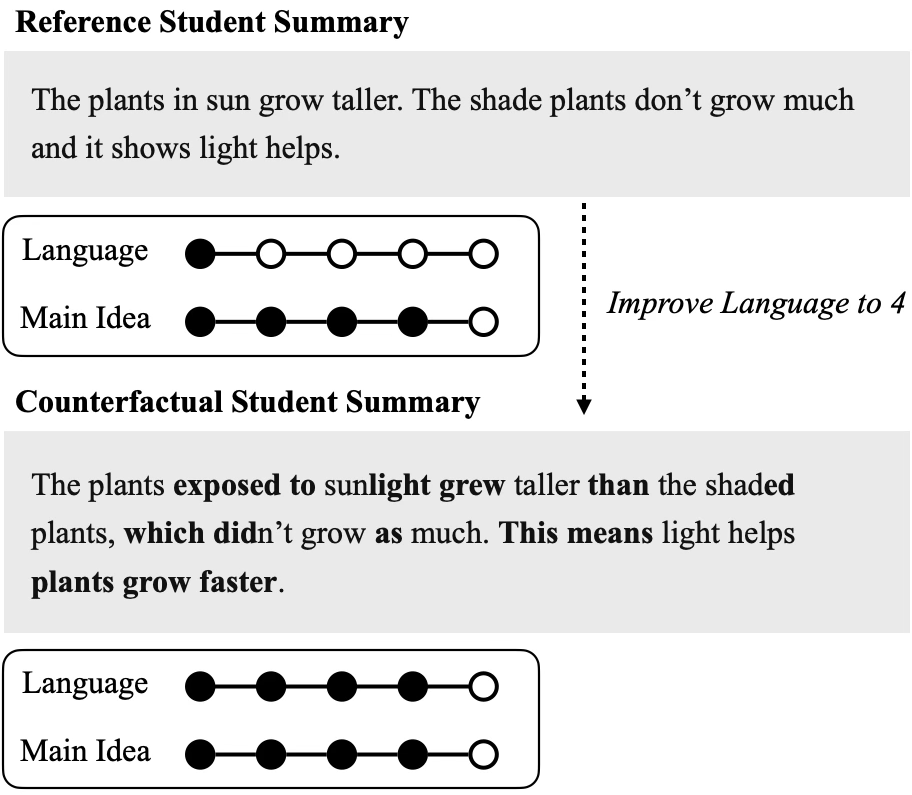}
    \caption{Overview of counterfactual student writing generation. Our approach produces a revised version of student work that improves a specific rubric criterion while remaining similar to the original.}
    \label{fig:placeholder}
    \vspace{-.4cm}
\end{figure}

While personalized counterfactual student work can be pedagogically helpful, manually crafting it can be labor-intensive and hard to scale. However, when student work is textual, recent advances in natural language processing, particularly Large Language Models (LLMs), provide opportunities for automation. Recent work leverages LLMs to generate counterfactual examples for a wide range of purposes, such as story rewriting \cite{chen_unsupervised_2022}, augmenting training data \cite{qiu_paircfr_2024, zhang_dually_2025}, or enhancing model interpretability \cite{ross-etal-2021-explaining}. These findings show the promise of using LLMs to automate the generation of counterfactual student work.

Existing LLM-based approaches to automated counterfactual generation must address a central design challenge: balancing the tradeoff of task performance and similarity to the original factual text. Since these two objectives are often in opposition, a variety of approaches exist to address this tradeoff. Some approaches optimize a language model for this joint objective \cite{jung_counterfactual_2022,hu_causal_2021,dathathri_plug_2019}, thereby encoding the trade-off in its weights. Another common pattern across approaches is to decompose the problem into identification and replacement steps, where one model identifies words of the factual text that should be revised and a second replaces them for the counterfactual task. The models used in each step are either LLMs trained for the specific step on a given task \cite{ross-etal-2021-explaining,treviso_crest_2023} or a large pretrained LLMs \cite{chen_disco_2023, dixit_core_2022, sachdeva_catfood_2024, wang_fitcf_2025}. This decomposition-based approach implicitly enforces similarity by the assumption that changing only the words with maximum influence will minimize the total number of words changed, thereby enforcing similarity by construction. 

While existing approaches provide promising directions towards generating counterfactual student work, significant work remains to make them adoptable in real classrooms \cite{filighera_towards_2022}. One limitation is the tight coupling between task alignment and similarity control: many approaches rely on training task-specific classifiers or control codes, which jointly define what constitutes a valid counterfactual and how to generate them. As a result, adapting to a new task often requires retraining or system redesign, which limits generalizability. A second limitation is that most existing works frame counterfactual generation in terms of nominal labels, such as sentiment, topic, or toxicity. In these settings, success is defined by crossing a decision boundary since counterfactuals are evaluated by the resulting change in label. 
In education, however, student work is often scored in an ordinal way. Therefore, counterfactual work needs to reflect nuanced adjustments in student work that lead to continuously improved quality.

\textbf{Contributions} In this paper, we propose the Gumbel Machine, a modular approach for counterfactual generation that decouples similarity control from counterfactual task alignment. We first introduce $\beta$-Hindsight control, a model-agnostic, inference-time controlled decoding mechanism that recovers and reuses the stochastic state implied by a reference to induce similarity during generation. Furthermore, we detail how to combine this mechanism with a preference optimization strategy to improve an LLM's adherence to a rubric. Together, these components form our counterfactual generation approach that enables both specific control over task validity and adjustable control over similarity to a reference.

We evaluate our combined approach on two real-world datasets of scored student-written essay summaries using an ordinal rubric, showing flexibility beyond standard nominal evaluation settings. We show that our approach yields counterfactuals that are more similar and valid compared to multiple baselines, on both open-weight and proprietary models.
In addition, we hire teachers to evaluate our approach and find that they consider our counterfactuals more similar to the original student work and, in such cases, prefer them for pedagogical use. 
Finally, we perform ablation studies on our approach to study the conditions under which the recovered stochastic state serves as an effective control signal.


\section{Background and Related Work}
\subsection{Gumbel-Max Trick}\label{sec:trick}

In our approach to generating counterfactual student work, we leverage the Gumbel-Max trick, which we briefly detail below. The Gumbel-Max trick is a procedure to sample from an arbitrary categorical distribution that involves adding noise sampled from a Gumbel distribution \cite{Maddison2014AS}. The samples from this procedure are shown to be equivalent to directly sampling from the categorical distribution, but the trick has the added benefit of decomposing the sampling into a reusable noise artifact that can be used to replay the randomness of the sampling process. 

\begin{figure*}[t]
    \centering
    \includegraphics[width=0.95\linewidth]{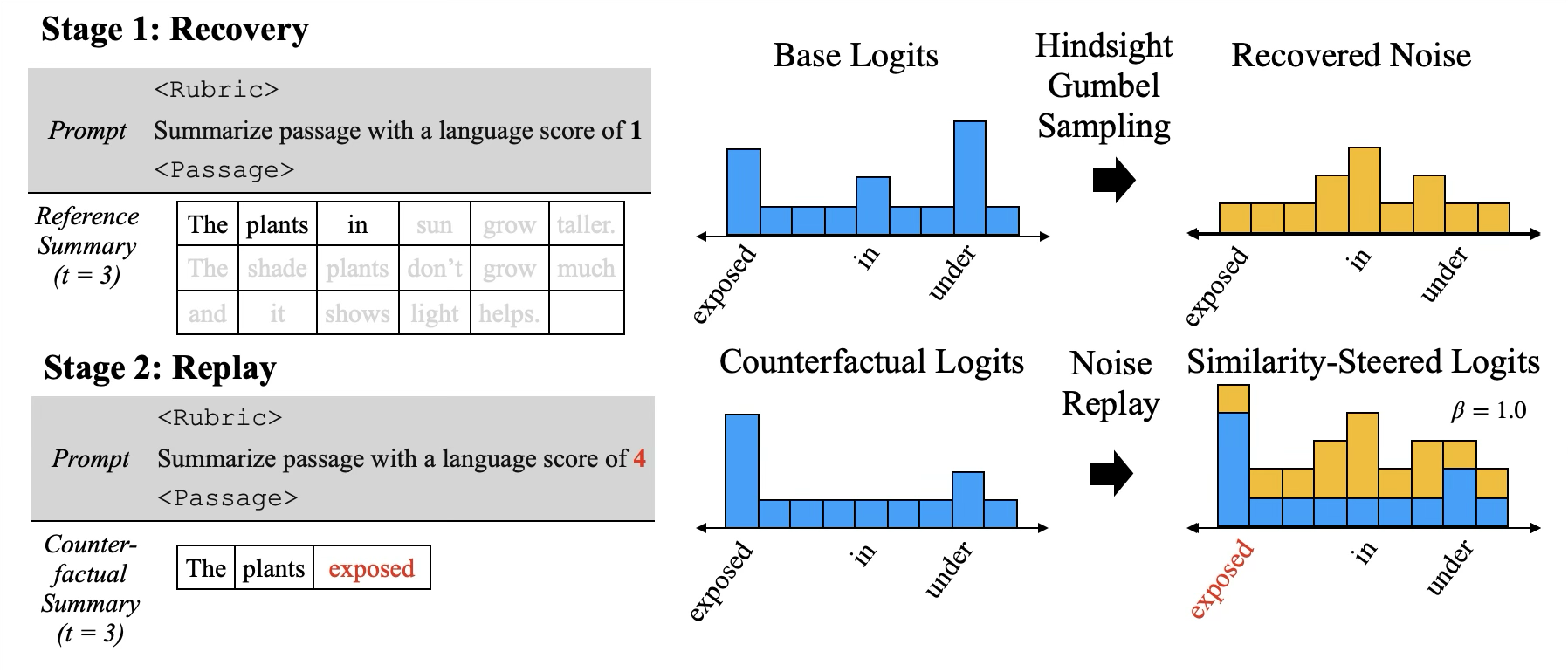}
    \vspace{-.4cm}
    \caption{Overview of the Gumbel Machine. In Stage 1, we recover noise that would have produced the observed tokens. In Stage 2, we intervene on the prompt with a new target score and replay the noise, steering generation toward a modified score while preserving similarity to the reference. In this example, ``exposed'' is selected, while the original token ``in,'' is a close alternative. A tunable $\beta$ scales the influence of noise-based similarity on replay.}
    \label{fig:placeholder}
    \vspace{-.3cm}
\end{figure*}

An LLM uses the softmax to sample tokens:
$$P(Y = v) = \frac{\exp(\ell_v)}{\sum_{v' \in V} \exp(\ell_{v'})}.$$
Here, $Y$ is a random variable for the next output token; $V$ is the model's vocabulary, the set of all tokens the model is trained to output; and $\ell_v$ are the \emph{logits} for each token $v \in V$. 
%
Since we express $Y$ in terms of logits, we can use the Gumbel-Max trick to sample the next token instead of sampling from $P(Y)$ directly. We draw a noise sample from the Gumbel distribution for each token and add it to the logits. We then take the maximum perturbed logit index and find the corresponding token: 
$$ y = \arg\max_{v \in V} (\ell_v + g_v).$$
Here, $g_v \sim \text{Gumbel}(0,1)$ is an independent draw from a standard Gumbel distribution. This simple procedure is mathematically equivalent to directly sampling from $P(Y=v)$. Moreover, the Gumbel noise values can be stored to preserve the random state of the sampling process.

In our approach, we use the recovered Gumbel noise for counterfactual generation, as it enables us to keep the exogenous component of sampling fixed after applying an intervention to a language model. Concretely, we record the sampled Gumbel values, update the LLM input, and then replay the Gumbel-Max trick with the recorded ``factual'' noise values on the counterfactual logits.

\subsection{Hindsight Gumbel Sampling}

The Gumbel-Max trick provides a mechanism for preserving and replaying the random noise used in discrete random processes. However, in practice, we often do not have access to the random process and can only observe samples. For example, the random process that generates student writing is controlled by the students themselves, not a computational system that produces logits. Therefore, we need to approximate the random process and recover the noise implied by the observations. \citet{ravfogel2025gumbel} refers to the task of recovering noise from a proxy as ``Hindsight Gumbel Sampling.'' In their work, the authors use an LLM as a proxy for the textual data generation process and outline an algorithmic procedure to recover Gumbel noise implied by data. We build on this foundation and demonstrate that recovered noise can serve as a tunable similarity-control mechanism. 

\subsection{Related Work}

\paragraph{Counterfactual Generation.} Counterfactual text generation has been widely studied, within NLP, for applications for a variety of tasks including sentiment analysis~\cite{kaushik_learning_2020, madaan_generate_2021}, story rewriting~\cite{chen_unsupervised_2022}, question answering~\cite{sachdeva_catfood_2024, paranjape_retrieval-guided_2022}, relation extraction~\cite{miao_episodic_2024}, and domain adaptation~\cite{calderon_docogen_2022, wu_polyjuice_2021}. Recent work has also explored applications of LLMs for counterfactual generation, particularly for data augmentation and model explainability \cite{goethals_what_2025, zhang_dually_2025,wang_fitcf_2025,gat_faithful_2023, wen_autocad_2022}.

\paragraph{Counterfactuals in Education.} Extensive work has been done in education to provide feedback for short answer writing~\cite{filighera_your_2022, mcnichols_can_2024, scarlatos_improving_2024}. Closely related to counterfactual generation is the task of paraphrase generation ~\cite{ouahrani_paraphrase_2024}, but existing work on counterfactual generation remains limited, highlighting the need for further development in this domain.  



\paragraph{Gumbel Noise for LLM Generation.} Recent work has explored the use of Gumbel noise for consistency during LLM generation \cite{de_mijolla_waste_2025}. Subsequent work explores the application of Gumbel noise to counterfactual generation for analyzing language model bias \cite{chatzi2025counterfactual}. \citet{ravfogel2025gumbel} introduces a procedure for reference-based recovery and replay of Gumbel noise for counterfactual generation. 

Our approach builds on this line of work and introduces several key innovations. First, this work demonstrates that the recovered Gumbel Noise can serve as a tunable reference-similarity control signal during generation time. Second, our work is the first to extensively evaluate the quality of the Gumbel noise-generated counterfactuals for an applied downstream task. Finally, we introduce a practical variant of hindsight Gumbel sampling that does not require computing the normalization constant.

\section{Methodology}
In this section, we formalize the task of counterfactual generation and our approach to this task. 

\subsection{Counterfactual Generation Task}
In our setting, we treat both factual and counterfactual outputs as token sequences generated by an LLM in response to an instruction prompt. Formally, we assume a pretrained autoregressive LLM $f_{\mathbf{\theta}}: \mathbf{x} \rightarrow \mathbf{y}$, parameterized by weights $\mathbf{\theta}$. We define $\mathbf{x}={( x_1 \dots x_N)}$ as a sequence of tokens forming the instruction (prompt) to the model, and $\mathbf{y}={( y_1 \dots y_M)}$ as the output tokens sampled from the model. We define a counterfactual as an alternative output $f(\mathbf{x'})_{\mathbf{\theta}}=\mathbf{y}'$ obtained after an intervention to the input prompt $\mathbf{x}'$. 

In counterfactual generation, our goal is to create an output that reflects a category change intervention in the input while remaining similar to a factual reference. Formally, given a reference $\mathbf{y}_r$ in category $z$, we need to output $\mathbf{y}'$ in target category $z'$, where $z \neq z' \in \{0,\dots ,k\}$ for a task with $k$ distinct categories. 
In tasks such as counterfactual topic generation, $z$ is interpreted as a nominal label, but in our counterfactual student work generation task, we interpret $z$ as an ordinal score. To determine a category for an arbitrary output, we introduce a scoring model $s: \mathbf{y} \rightarrow \hat{z}$ to estimate a score for an arbitrary output, and we say a counterfactual is \emph{valid} when $s(\mathbf{y}')=z'$.

\subsection{$\beta$-Hindsight Similarity Control}


We now detail our inference-time similarity control approach for counterfactual generation, outlined in Algorithm~\ref{alg:gumbel-cf-short}. There are two phases: recovering the noise under a base LLM and replaying the noise values to sample from an intervened LLM.

To recover the Gumbel noise that corresponds to each token of a reference, for each autoregressive timestep $t$, we derive a noise vector $\mathbf{g}_t \in \mathbb{R}^V$ that satisfies the condition: \begin{equation} \label{max-condition}
    \ell[{y_t}] + \mathbf{g}_t[{y_t}] > \ell[v] + \mathbf{g}_t[v]\  \forall v \in V, v \ne y_t.
\end{equation}
Here, $V$ is the LLM's vocabulary, $y_t$ indexes the factual reference token at timestep $t$, and $\ell[{y_t}]$ is the associated logit. We omit the subscript $r$ of the reference $y_r$ for clarity in time-based indexing. This condition ensures that the observed factual token $y_t$ is selected during the Gumbel-Max prediction.

In order to satisfy Equation~\ref{max-condition}, we form a noise vector by first sampling the target noise from a lower-truncated Gumbel distribution. 
$$
\mathrm{LowerTruncG}(\tau_{min}) \coloneqq P\bigl(G \mid G > \tau_{min}\bigr).
$$ 
Here, $\tau_{min}$ denotes the lower bound of a truncated standard Gumbel distribution. We then compute $\tau_{min} = \ell_{max} - \ell[{y_t}]$, the difference between the ground truth token logit and the maximum observed logit, so that the augmented logit for the reference token will be greater than the logit for any other token at each timestep.

\input{algo_v2_short}

The sampled target noise establishes a ceiling, $T_{t} = \ell[{y_t}] + g[y_t]$, the new logit value for the reference token. Subsequently, for all other tokens $v \neq y_t$, we sample their corresponding noise values from an upper-truncated Gumbel distribution, such that no resulting token logit exceeds the ceiling. 
$$
\mathrm{UpperTruncG}(\tau_{max}) \coloneqq P\bigl(G \mid G < \tau_{max}\bigr).
$$ 
Here, $\tau_{max}$ denotes the upper bound of a truncated standard Gumbel distribution. We compute $\tau_{max} = T_t - \ell[v]$, the difference between the ceiling logit value and the actual logit value, for each of the other tokens in the vocabulary, so that none of the resulting logit values would exceed the ceiling. To efficiently sample from the truncated Gumbel distributions, we perform inverse transform sampling \cite{devroye1986non}, mapping the uniform samples to the truncated support range before applying the inverse CDF. The recovered Gumbel noise values satisfy (\ref{max-condition}) by construction and represent a stochastic state that would have led the LLM to output the token in the reference, provided the prompt and preceding tokens.

Given the recovered noise values, we then replay the noise to generate the counterfactual output under a modified prompt as input. The replay procedure follows from the Gumbel-max trick, in which we additionally modulate the noise distribution with a scaling hyperparameter $\beta$. This hyperparameter controls the strength of similarity during sampling, with higher values inducing LLM outputs more similar to the reference. 

\subsection{Instruction Alignment for Task Validity} \label{train-setup}

We now detail our approach to encourage validity in generated counterfactual student work. We frame this validity alignment problem as an instruction-following task, in which we prompt a language model with explicit instructions on how to generate a valid counterfactual. In each prompt, we include a rubric detailing the criterion to change, the qualitative aspects of each possible ordinal score $z$, and the reference student work $\mathbf{y_r}$. We conclude the prompt with instructions for the model to generate a revised version $y'$ that adheres closely to the reference while reflecting a specified score $z'$. We include the prompts in Appendix~\ref{apx:prompts}.

We find that training with preference alignment substantially improves task validity. We first perform supervised fine-tuning (SFT) \cite{ouyang2022training} using examples derived from a dataset of $(\mathbf{y}, z)$ tuples, where $z$ is used in forming the prompt $\textbf{x}$ and $\textbf{y}$ is the target. During training, we use the same prompt format but exclude the reference and related adherence instructions. After conducting SFT, we further align the model to the counterfactual task using direct preference optimization (DPO) \cite{rafailov2023direct}. To construct preference pairs, we use student examples that achieve a target, ground-truth score on a given criterion as a preferred example, and those that receive an alternative score for the same criterion as a negative example. For each example in the training dataset, we randomly select one negative sample for each alternative score on the same prompt, yielding $k-1$ preference pairs per example. Together, instruction-based prompting, supervised fine-tuning, and preference alignment enable reliable task validity and are compatible with $\beta$-Hindsight similarity control.

\section{Experimental Setup}
In this section, we detail our experiments to evaluate the effectiveness of our approach.

\subsection{Datasets}
We evaluate our approach on two datasets of scored student writing samples. 
The first dataset is Common Lit Augmented Student Summary Evaluation (\textbf{CLASSE}), a corpus of student-authored summaries of short reading passages \cite{crossley-etal-2024-world}. The dataset contains 4,689 summaries of 101 unique passages authored by students in grades 3-12. 
The second dataset is Dataset for Rubric-based Essay Scoring on EFL Writing (\textbf{DREsS}), a corpus of student-authored responses to open-ended writing prompts \cite{yoo_dress_2025}. We use $\text{DREsS}_{\text{New}}$, which contains 2,279 essays from 65 essay writing prompts authored by university undergraduate students enrolled in an English as a foreign language (EFL) writing course. The student writings in both datasets are graded by experts with ordinal rubrics on multiple criteria for writing quality, 5 for CLASSE and 3 for DREsS. We further detail criterion definitions, token statistics, and scoring rubrics for the datasets in Appendix~\ref{apx:classe}.



\subsection{Metrics}
We measure two competing properties of the generated counterfactuals, i) similarity to a factual example and ii) task validity, following existing work on counterfactual generation \cite{wang-etal-2024-survey}. 

\paragraph{Similarity}
In our experiments, we measure the similarity between the reference student summary and the counterfactual student summary. We report the complement of the normalized character-level Levenshtein distance, i.e., one minus the normalized distance, averaged across all $N$ counterfactual summaries for a given setting:
$\text{Similarity}=\frac{1}{N}\sum_{i=1}^N \big(1 -\frac{\text{Lev}(y_i, y_i')}{\text{max}(\lVert y_i \rVert, \lVert y_i' \rVert)}\big).$
Here, $y_i$ is the reference summary and $y_i'$ the counterfactual summary. This metric outputs a score in $[0,1]$ where 1 indicates maximum similarity.

\paragraph{Validity}
A generated counterfactual is valid if its score, given by the scoring model, reflects the intended score for the specified criterion. Existing evaluation metrics for validity, such as flip rate \cite{ross-etal-2021-explaining}, are for nominal tasks (e.g., sentiment or topic classification). Since scores in our task are ordinal, we use average Quadratic Weighted Kappa (QWK) across scoring criteria between the intended score $z'$ and an estimated score from the scoring model $s(\mathbf{y}')=\hat{z}'$. QWK produces a score in $[-1,1]$, with high values indicating higher task validity. 

We use an LLM-as-a-judge \cite{li_generation_2025} system to compute validity using a scoring rubric. Specifically, we fine-tune \texttt{GPT-4.1-mini} for each criterion across both datasets. We evaluate the average QWK agreement across scoring criteria between this model and human scores for CLASSE and DREsS, which are 0.6242 and 0.6012, respectively. This agreement is very close to the 0.642 inter-rater agreement for CLASSE (DREsS is single-scored), indicating that our scoring model is a good proxy of human scoring. We detail experiments on multiple approaches for automated validity scoring in Appendix~\ref{apx:scoring}.




\begin{table}[t]
\centering
\setlength{\tabcolsep}{3pt}
\begin{tabular}{l  cc  cc}
\toprule
& \multicolumn{2}{c}{\textbf{CLASSE}} & \multicolumn{2}{c}{\textbf{DREsS}} \\
\cmidrule(lr){2-3} \cmidrule(lr){4-5}
\multirow{-2}{*}{\textbf{Approach}} & $\text{Sim}^\uparrow$ & $\text{Val}^\uparrow$ & $\text{Sim}^\uparrow$ & $\text{Val}^\uparrow$ \\
\midrule
ME 
    & $0.314$ & $0.183$
    & $0.305$ & $0.188$ \\
ICL
    & $0.308$ & $\underline{0.327}$
    & $0.330$ & $0.264$ \\
I\&R 
    & $\textbf{0.498}$ & $0.103$
    & $0.308$ & $0.325$ \\
GM (Ours)
    & $\underline{0.391}$ & $\textbf{0.668}^{*}$
    & $\textbf{0.477}^{*}$ & $\textbf{0.404}^{*}$ \\
\bottomrule
\end{tabular}
\caption{Comparing GM to other counterfactual generation methods on both datasets (ME - Minimal-Edit, ICL - In-Context Learning, I\&R - Identify-and-Replace, GM - Gumbel Machine).}
\label{tab:llama_results}
\vspace{-.3cm}
\end{table}

\subsection{Experimental Settings and Details}

In this section, we detail the settings and baseline approaches we use to evaluate our approach.

\paragraph{Counterfactual Generation Baselines.}

We compare our approach, \textbf{Gumbel Machine (GM)}, to three other LLM-based counterfactual text generation strategies.
%
\textbf{Minimal-Edit (ME)}: Following \citet{gat_faithful_2023}, we include in the prompt instructions for the LLM to minimally edit the reference as a zero-shot prompting approach.
\textbf{In-Context Learning (ICL)}: Following \citet{li_prompting_2024}, we further include in-context demonstrations of each score level in the prompt.
\textbf{Identify-and-Replace (I\&R)}: Following \citet{nguyen_llms_2024}, we instruct the LLM to perform a series of steps to (i) identify word spans leading to the original score, (ii) change the identified spans to reflect the desired score, and (iii) construct a new summary with the modified spans. These instructions resemble a ``Chain-of-Thought (CoT)'' for our task \cite{wei2022chain}.

\begin{table}[t]
\small
\centering
\setlength{\tabcolsep}{4pt}
\begin{tabular}{ll  cc  cc}
\toprule
& & \multicolumn{2}{c}{\textbf{CLASSE}} & \multicolumn{2}{c}{\textbf{DREsS}} \\
\cmidrule(lr){3-4} \cmidrule(lr){5-6}
\multirow{-2}{*}{\textbf{Model}} & \multirow{-2}{*}{\textbf{Method}} & $\text{Sim}^\uparrow$ & $\text{Val}^\uparrow$ & $\text{Sim}^\uparrow$ & $\text{Val}^\uparrow$ \\
\midrule
\multirow{3}{*}{Haiku}
  & ME 
    & $0.280$ & $0.144$
    & $0.312$ & $0.093$ \\
  & ICL
    & $0.281$ & $0.312$
    & $0.314$ & $0.125$ \\
  & I\&R 
    & $0.451$ & $0.388$
    & $0.384$ & $\underline{0.441}$ \\
\midrule
\multirow{3}{*}{Sonnet}
  & ME 
    & $0.376$ & $0.344$
    & $0.375$ & $0.439$ \\
  & ICL
    & $0.359$ & $\underline{0.604}$
    & $0.418$ & $\textbf{0.484}$ \\
  & I\&R 
    & $\underline{0.554}$ & $0.434$
    & $\underline{0.477}$ & $0.405$ \\
\midrule
\multirow{1}{*}{Llama}
 & GM (Ours)
    & $0.391$ & $\textbf{0.668}^{*}$
    & $\underline{0.477}$ & $0.404$ \\
\midrule
\multirow{1}{*}{Qwen}
  & GM (Ours)
    & $\textbf{0.575}^{*}$ & $0.566$
    & $\textbf{0.593}^{*}$ & $0.315$ \\
\bottomrule
\end{tabular}
\caption{Comparing counterfactual LLM approaches across proprietary and open-weight models (ME - Minimal-Edit, ICL - In-Context Learning, I\&R - Identify-and-Replace, GM - Gumbel Machine). GM translates across open-weight models and has comparable performance to larger proprietary models.}
\label{tab:gm_vs_closed}
\end{table}

\paragraph{LLMs.}
We experiment with two open-weight models and two proprietary ones. 
We use \texttt{LLaMA-3.1-8B-Instruct} as our primary open-weight model due to its strong instruction-following behavior and wide community adoption \cite{grattafiori2024llama3herdmodels}, and \texttt{Qwen3-4B-Instruct} \cite{yang2025qwen3technicalreport}, which uses a different training pipeline, to evaluate whether our approach generalizes across different models. We also use \texttt{Claude Haiku 4.5} \cite{anthropic2025haiku45}, a lightweight prompting-only baseline at a comparable capability tier to our open-weight models, and \texttt{Claude Sonnet 4.6} \cite{anthropic2025sonnet46}, a larger and more capable proprietary model. This model choice for proprietary, prompting-based baselines mitigates bias when an LLM evaluator and a generator share a model family \cite{panickssery2024llmevaluators}, since our scoring model is based on \texttt{GPT-4.1-mini}.



\begin{table*}[ht]
\centering
\small
\begin{tabular}{l c c c c c c c}
\toprule
& \multicolumn{3}{c}{Per-side Likert (1--3)} & \multicolumn{3}{c}{Pairwise} & Conditional on sim=GM \\
\cmidrule(lr){2-4} \cmidrule(lr){5-7} \cmidrule(lr){8-8}
Dimension & GM & ME & Wilcoxon $p$ & GM/ME/Tie & GM win\% & Binomial $p$ & GM win\% (GM/ME/Tie) \\
\midrule
Score Improvement       & 2.58 & 2.64 & 0.41 & 83/82/35 & 50\% & 1.00 & \textbf{72\%} (44/17/3) \\
\textbf{Similarity} & \textbf{1.77} & 1.67 & \textbf{0.03} & \textbf{64/39/97} & \textbf{62\%} & \textbf{0.02} & --- \\
Utility      & 2.43 & 2.46 & 0.75 & 88/78/34 & 53\% & 0.49 & \textbf{75\%} (47/16/1) \\
\bottomrule
\end{tabular}
\caption{Human evaluation results across four evaluators comparing GM (Gumber Machine) to the ME (Minimal-Edit) baseline. Evaluators give GM counterfactuals higher similarity ratings in comparative and standalone evaluations. In those cases, generated feedback is also rated more useful. \textbf{Bold} indicates statistical significance.}
\label{tab:human_eval}
\end{table*}

\paragraph{Data Sampling Strategy.}
We generate counterfactuals for a stratified subset of score changes sampled from a held-out validation set for each dataset. Specifically, we perform an 80-20 train-validation split and sample uniformly from the validation split across all ordered score transitions (e.g., $4 \rightarrow 1$, $2 \rightarrow 3$), capping the total at 400 examples per rubric criterion and balancing across student writing tasks. This setting results in 2,000 and 1,200 total generations per experimental datapoint for CLASSE and DREsS, respectively. This strategy ensures balanced representation of both small and large counterfactual shifts in scores.


\section{Results, Analysis, Discussion}

We now detail experimental results and perform both human evaluation and ablation studies to verify the effectiveness of our approach.

\subsection{Counterfactual Approach Comparisons}

In Table~\ref{tab:llama_results}, we report the results of our experiments comparing GM to baseline LLM-based counterfactual approaches. In this experiment, we run all settings on Llama-3 and set $\beta=0.1$ and $\beta=0.05$ for CLASSE and DREsS, respectively, based on tuning experiments reported in Appendix~\ref{apx:tuning}. On CLASSE, we observe that our approach noticeably improves validity across both datasets and yields higher similarity scores. While the I\&R approach shows a higher similarity score than GM, its validity is the lowest among the approaches. This result suggests that I\&R produces counterfactuals that are more similar to the reference but not accurate with respect to the rubric, while the counterfactuals produced by GM are more accurate and similar. On DresS, we observe that GM outperforms all baselines on both metrics, but do not see as large of an improvement on validity. This observation suggests that GM can produce more rubric-aligned counterfactuals when the writing samples are more homogeneous, e.g., for summarization, compared to open-ended writing. 

\subsection{Cross-LLM Comparisons}

In Table~\ref{tab:gm_vs_closed}, we report results comparing our approach to other models, Qwen-3 and Claude (prompting-only). 
On CLASSE, Llama-3-based GM outperforms the strongest Sonnet-based validity baseline (ICL) on both metrics; Qwen-3-based GM outperforms Sonnet's strongest similarity baseline (Identify-and-Replace) across both metrics. On DREsS, GM improves on similarity compared to proprietary models, but does not reach the same levels of validity in some settings. This result suggests that for longer open-ended writing tasks, the capacity of a larger, more capable base LLM helps. 
Moreover, the result indicates that combining our approach with a larger open-weight model may yield further improvements, but at the expense of higher computational costs. Overall, these results suggest that our GM approach is generalizable across open-weight model families; more importantly, we obtain counterfactuals that are either better than or comparable to those generated by a much larger proprietary model.

\subsection{Human Evaluation}

We conduct an IRB-approved human evaluation study to evaluate the effectiveness of our approach and the quality of the generated counterfactual student writing. We employ four evaluators via Upwork \cite{upwork}, all of whom have experience teaching English in the United States across elementary, middle school, high school levels.

We focus our evaluation on student summaries from the CLASSE dataset, since the shorter, student-written summaries in CLASSE are easier to evaluate at scale than the DREsS essays. We show each evaluator 50 student writing summaries and two counterfactual rewrites of each summary. We then ask them to evaluate each pair of rewritten summaries and compare them across three dimensions: improvement relative to the scoring criterion, similarity to the original summary, and overall pedagogical utility. We select 10 writing samples for each of the 5 criteria and stratify sampling by the original student score on each criterion to ensure that the evaluation includes student writing at different quality levels. We generate rewrites with the targeted score set to the maximum score of 4, sampling from both the GM and the ME baselines. In this setting, both approaches generate rewrites with similar validity, yielding average scores of $3.64/4$ and $3.68/4$, respectively. Therefore, our evaluation setting enables a fair comparison between the two approaches that generate different rewrites but at the same level of validity. To prevent positional bias, we randomize the ordering of rewrites from both approaches.

We report the results of this evaluation in Table~\ref{tab:human_eval}. We observe that evaluators rate GM outputs as more similar to the reference student writing in 62\% of contested cases ($p=0.018$), and give a significantly higher aggregate similarity rating ($p=0.033$). Moreover, our automated similarity metric has moderate yet statistically significant correlation with average teacher similarity ratings ($\rho=0.48$, $p<0.001$). We find no significant difference between the approaches in terms of score improvement (as expected) and utility. However, we observe that when the evaluator judges the GM rewrite as more similar, they also prefer GM on utility in 75\% of contested cases ($\chi^2(4)=57.8$, $p<0.001$). 
These results highlight reference similarity as an important factor in creating pedagogically useful feedback for students and demonstrate that fine-grained control techniques beyond prompt design are directionally necessary towards creating automated rewriting feedback systems that are faithful to student writing preferences. We report further details and results of our teacher evaluation in Appendix~\ref{sec:apdx-human-eval}.


\begin{table}[t]
\small
\centering
\setlength{\tabcolsep}{3pt}
\scalebox{1}{
\begin{tabular}{ccc | cc | cc}
\toprule
& & & \multicolumn{2}{c|}{\textbf{CLASSE}} & \multicolumn{2}{c}{\textbf{DREsS}} \\
\cmidrule(lr){4-5} \cmidrule(lr){6-7}
\multirow{-2}{*}{\textbf{$\beta$-Hindsight}} & \multirow{-2}{*}{\textbf{SFT}} & \multirow{-2}{*}{\textbf{DPO}} & $\text{Sim}^\uparrow$ & $\text{Val}^\uparrow$ & $\text{Sim}^\uparrow$ & $\text{Val}^\uparrow$ \\
\midrule
$-$ & $-$ & $-$
    & $0.314$ & $0.183$
    & $0.305$ & $0.188$ \\
$-$ & \checkmark & $-$
    & $0.382$ & $0.151$
    & $0.290$ & $0.102$ \\
$-$ & \checkmark & \checkmark
    & $0.324$ & $\textbf{0.688}$
    & $0.224$ & $\textbf{0.608}$ \\
\midrule
\checkmark & $-$ & $-$
    & $\underline{0.560}$ & $0.168$
    & $0.402$ & $0.148$ \\
\checkmark & \checkmark & $-$
    & $\textbf{0.795}$ & $\text{-}0.026$
    & $\textbf{0.733}$ & $\text{-}0.005$ \\
\checkmark & \checkmark & \checkmark
    & $0.391$ & $\underline{0.668}$
    & $\underline{0.477}$ & $\underline{0.404}$ \\
\bottomrule
\end{tabular}}
\caption{Ablation study on the components of GM. We observe that SFT alone drastically improves similarity at the cost of validity and that DPO is necessary to realize counterfactual results that balance both axes.}
\label{tab:ablation_alignment}
\end{table}

\subsection{Ablation Studies}\label{sec:foresight}

In Table~\ref{tab:ablation_alignment}, we report the results of an ablation on the training and logit control components of GM, running each configuration on Llama-3. We observe a distinctive trend in the introduction of each component. With $\beta$-Hindsight decoding alone and no training, we observe improved similarity at marginally lower validity. This result suggests that recovered Gumbel noise can steer the sampling trajectory toward the reference but does not induce rubric-adherent counterfactuals. Training the model with SFT on student writing amplifies the noise's control influence, resulting in a large jump in similarity while validity collapses to near zero. This result suggests that SFT is necessary to adapt a model to student writing style, which may not be statistically likely for a pre-trained model. 
Notably, adding DPO after SFT significantly improves the model's capability on this task. This result suggests that by contrasting generations with different scores, DPO helps the LLM with controllable generation, while $\beta$-Hindsight remains key to maintaining similarity to original student writing, which prompting alone cannot achieve. All three components are therefore necessary to produce valid counterfactuals that are similar to the reference. We report the results of an additional ablation study on the prompt configuration in Appendix~\ref{apx:prompt-ablate}.

\section{Conclusions and Future Work}
We present the Gumbel Machine, a modular approach to counterfactual generation that leverages preference alignment and replayed random state with $\beta$-Hindsight control.
We evaluate our approach on two datasets of student-written summaries and demonstrate that it effectively generates counterfactual student work that is both rubric-valid and similar to the original. This work demonstrates that recovered generative randomness can serve as a tunable control signal with immediate application.

Our work outlines future directions that include: evaluating this approach across other domains and generation tasks, assessing the pedagogical effectiveness of generated counterfactuals relative to conventional writing feedback, and further investigation into the necessary and sufficient conditions for recovering noise that is effective for control.

\newpage
\section*{Limitations}

While our results suggest that recovered Gumbel noise can serve as a practical similarity control signal for counterfactual student writing, there are limitations that constrain the interpretation and generalizability of our findings. 

First, our recovery-and-replay procedure assumes that the data-generating process for student writing can be approximated by the autoregressive Gumbel-Max structural causal model (SCM) induced by the proxy LLM. As prior work has noted, the Gumbel-Max SCM is one of several parameterizations of categorical sampling and is a modeling choice rather than a discovered structure \citep{chatzi2025counterfactual, benz2025evaluation}. Real student writing also involves planning, revision, and non-local edits, so the strict autoregressive assumption is at best an approximation. Our results should therefore be read as counterfactuals \emph{under a proxy autoregressive SCM}, rather than as causal counterfactuals over the true student-writing process. Our ablation study supports this interpretation where we observe that a pre-trained LLM, optimized for well-written summaries, assigns low likelihood to student-authored ones (which can contain rubric-relevant errors), so the noise it produces is residual-like rather than meaningful. However, upon training the LLM on student data, the noise becomes a much more effective control signal, and thus a better, but still limited, proxy model of student writing.

Second, our approach assumes access to carefully engineered prompts that specify the rubric, request adherence to a reference, and clearly define the target score change. As shown in an ablation study, reference conditioning is a key enabling condition for effective control, suggesting sensitivity to prompt variation that does not preserve this structure. In addition, we train separate alignment adapters per rubric criterion, which improves controllability in our experiments but may not scale cleanly to richer, multi-criterion feedback without more advanced multitask or compositional alignment strategies.

Third, while our approach is compared across LLMs of both proprietary and open-weight types, our comparative baselines are primarily prompting-based. While these baselines are representative of recent research on counterfactual generation approaches, their comparison with the Gumbel Machine is somewhat mismatched, as they do not directly modify output logits. While a body of prior research has explored logit manipulation techniques for counterfactual generation, these techniques do not translate to ordinal task setups as they are designed for nominal tasks and would require significant modification to adapt to our task, challenging direct comparison \cite{krause2021gedi, yang2021fudge, dathathri_plug_2019}. To mitigate this evaluation gap, we provide a comparison to Vocab Bias, a direct logit manipulation technique in Appendix~\ref{apx:tuning}.

Finally, our primary validity measurements rely on a fine-tuned GPT-4.1-mini judge (Table~\ref{tab:human_eval}) to enable scalable evaluation. While comparisons to ground-truth inter-rater agreement are promising, the automated judge may introduce systematic biases and may not fully capture pedagogical usefulness. Likewise, similarity is measured using normalized character-level Levenshtein distance, which is common in related literature but is sensitive to surface edits and not semantics-aware. Our human evaluation finds that this metric correlates moderately with teacher judgment of similarity, but interpretation should still be made with caution. Furthermore, our approach has not been directly studied with students, so teacher useful ratings are only an approximation for how well this approach would translate to meaningful learning improvements.

\section*{Ethical Considerations}
Counterfactual student writing, even when intended as illustrative feedback, risks being perceived as an “ideal rewrite” rather than a pedagogical example, which could unintentionally discourage student agency or mask the student’s original voice. If deployed without appropriate framing, such systems may encourage over-reliance on AI-generated revisions, shifting learning from skill development toward imitation. These risks underscore the importance of positioning counterfactuals as illustrative demonstrations rather than authoritative corrections in classroom environments.

On fairness, bias, and accountability, both the generation and evaluation components of our system rely on LLMs that may encode biases related to language background, dialect, or writing style, potentially advantaging certain student populations over others. Automated scoring models used to assess validity may further amplify these biases, especially when subtle stylistic features are conflated with quality. Additionally, using student writing as input raises privacy concerns, particularly in classroom deployments involving minors. Any real-world use of this approach should therefore include safeguards such as bias audits, human oversight, transparency about AI involvement, and follow local regulations regarding the use of student data.

\clearpage

\bibliography{custom}

\clearpage
\appendix

\section{Appendix}\label{sec:appendix}


\subsection{Per-Criterion Results} \label{apx:tradeoffs}

\begin{table*}[!htbp]
\centering
\setlength{\tabcolsep}{3pt}
\scalebox{.85}{
\begin{tabular}{ll  cc  cc  cc  cc  cc}
\toprule
\textbf{Model} & \textbf{Approach}
  & $\text{Sim}^\uparrow$ & $\text{Val}^\uparrow$
  & $\text{Sim}^\uparrow$ & $\text{Val}^\uparrow$
  & $\text{Sim}^\uparrow$ & $\text{Val}^\uparrow$
  & $\text{Sim}^\uparrow$ & $\text{Val}^\uparrow$
  & $\text{Sim}^\uparrow$ & $\text{Val}^\uparrow$ \\
\midrule
\multicolumn{2}{l}{\textit{CLASSE}}
  & \multicolumn{2}{c}{\textbf{Main Idea}}
  & \multicolumn{2}{c}{\textbf{Details}}
  & \multicolumn{2}{c}{\textbf{Organization}}
  & \multicolumn{2}{c}{\textbf{Wording}}
  & \multicolumn{2}{c}{\textbf{Language}} \\
\midrule
\multirow{3}{*}{Haiku}
  & ME       & 0.285 & 0.064 & 0.282 & 0.243 & 0.279 & 0.015 & 0.266 & 0.341 & 0.289 & 0.056 \\
  & ICL      & 0.274 & $\textbf{0.365}$ & 0.268 & 0.454 & 0.279 & 0.158 & 0.292 & $\textbf{0.513}$ & 0.294 & 0.072 \\
  & I\&R     & $\textbf{0.482}$ & 0.263 & $\textbf{0.402}$ & $\textbf{0.651}$ & $\textbf{0.501}$ & $\textbf{0.177}$ & $\textbf{0.407}$ & 0.486 & $\textbf{0.461}$ & $\textbf{0.365}$ \\
\cmidrule(l){2-12}
\multirow{3}{*}{Sonnet}
  & ME       & 0.450 & 0.139 & 0.344 & 0.599 & 0.348 & 0.114 & 0.362 & 0.651 & 0.377 & 0.216 \\
  & ICL      & 0.377 & $\textbf{0.577}$ & 0.320 & 0.711 & 0.362 & $\textbf{0.496}$ & 0.380 & $\textbf{0.804}$ & 0.356 & $\textbf{0.432}$ \\
  & I\&R     & $\textbf{0.646}$ & 0.410 & $\textbf{0.520}$ & $\textbf{0.766}$ & $\textbf{0.562}$ & 0.201 & $\textbf{0.524}$ & 0.419 & $\textbf{0.516}$ & 0.373 \\
\cmidrule(l){2-12}
\multirow{4}{*}{Llama}
  & ME       & 0.305 & 0.165 & 0.315 & 0.224 & 0.299 & 0.265 & 0.341 & 0.154 & 0.310 & 0.107 \\
  & ICL      & 0.288 & 0.272 & 0.308 & 0.381 & 0.305 & 0.516 & 0.320 & 0.182 & 0.320 & 0.282 \\
  & I\&R     & $\textbf{0.528}$ & 0.117 & $\textbf{0.440}$ & 0.096 & $\textbf{0.529}$ & 0.112 & $\textbf{0.498}$ & 0.092 & $\textbf{0.494}$ & 0.099 \\
  & GM (Ours) & 0.364 & $\textbf{0.820}$ & 0.334 & $\textbf{0.780}$ & 0.387 & $\textbf{0.769}$ & 0.412 & $\textbf{0.666}$ & 0.458 & $\textbf{0.306}$ \\
\cmidrule(l){2-12}
\multirow{4}{*}{Qwen}
  & ME       & $\textbf{0.648}$ & 0.163 & $\textbf{0.667}$ & 0.247 & 0.579 & 0.239 & 0.558 & 0.281 & 0.432 & 0.230 \\
  & ICL      & 0.349 & 0.190 & 0.347 & 0.248 & 0.359 & 0.156 & $\textbf{0.620}$ & 0.217 & 0.326 & 0.137 \\
  & I\&R     & 0.524 & 0.282 & 0.403 & 0.540 & 0.474 & 0.328 & 0.388 & $\textbf{0.385}$ & 0.450 & 0.286 \\
  & GM (Ours) & 0.506 & $\textbf{0.729}$ & 0.455 & $\textbf{0.761}$ & $\textbf{0.624}$ & $\textbf{0.483}$ & 0.615 & 0.312 & $\textbf{0.677}$ & $\textbf{0.544}$ \\
\midrule
\multicolumn{2}{l}{\textit{DREsS}}
  & \multicolumn{2}{c}{\textbf{Content}}
  & \multicolumn{2}{c}{\textbf{Language}}
  & \multicolumn{2}{c}{\textbf{Organization}}
  & \multicolumn{4}{c}{} \\
\midrule
\multirow{3}{*}{Haiku}
  & ME       & 0.308 & 0.105 & 0.337 & 0.123 & 0.292 & 0.052 & & & & \\
  & ICL      & 0.307 & 0.170 & 0.328 & 0.097 & 0.307 & 0.109 & & & & \\
  & I\&R     & $\textbf{0.340}$ & $\textbf{0.566}$ & $\textbf{0.460}$ & $\textbf{0.216}$ & $\textbf{0.352}$ & $\textbf{0.541}$ & & & & \\
\cmidrule(l){2-12}
\multirow{3}{*}{Sonnet}
  & ME       & 0.332 & 0.544 & 0.431 & $\textbf{0.411}$ & 0.363 & 0.363 & & & & \\
  & ICL      & 0.372 & $\textbf{0.556}$ & 0.502 & 0.386 & 0.380 & 0.509 & & & & \\
  & I\&R     & $\textbf{0.438}$ & 0.435 & $\textbf{0.571}$ & 0.212 & $\textbf{0.426}$ & $\textbf{0.566}$ & & & & \\
\cmidrule(l){2-12}
\multirow{4}{*}{Llama}
  & ME       & 0.312 & 0.323 & 0.307 & 0.143 & 0.295 & 0.097 & & & & \\
  & ICL      & 0.357 & 0.327 & 0.320 & 0.239 & 0.312 & 0.227 & & & & \\
  & I\&R     & 0.296 & $\textbf{0.489}$ & 0.322 & 0.116 & 0.307 & 0.370 & & & & \\
  & GM (Ours) & $\textbf{0.637}$ & 0.307 & $\textbf{0.498}$ & $\textbf{0.371}$ & $\textbf{0.344}$ & $\textbf{0.507}$ & & & & \\
\cmidrule(l){2-12}
\multirow{4}{*}{Qwen}
  & ME       & 0.298 & 0.266 & 0.305 & 0.195 & 0.287 & 0.128 & & & & \\
  & ICL      & 0.301 & 0.235 & 0.388 & 0.104 & 0.324 & 0.100 & & & & \\
  & I\&R     & 0.248 & $\textbf{0.718}$ & 0.337 & $\textbf{0.312}$ & 0.351 & $\textbf{0.582}$ & & & & \\
  & GM (Ours) & $\textbf{0.511}$ & 0.414 & $\textbf{0.604}$ & 0.200 & $\textbf{0.707}$ & 0.248 & & & & \\
\bottomrule
\end{tabular}}
\caption{Per-criterion breakdown comparing counterfactual approaches (ME - Minimal-Edit, ICL - In-Context Learning, I\&R - Identify-and-Replace, GM - Gumbel Machine) on Similarity and Validity on the CLASSE (top) and DREsS (bottom) datasets. Bold marks the best value per model within each (criterion, metric) cell. }
\label{tab:appendix_per_criterion}
\end{table*}
We report a per-criterion expansion of the aggregate similarity and validity measurements reported for Llama-3 and Qwen3 in 
Table~\ref{tab:appendix_per_criterion}. We observe that the within-model similarity-validity ordering is consistent across criteria for both open-weight backbones where GM tends to maximize validity with minimal tradeoff for similarity. The clearest exception is the CLASSE ``Wording'' criterion, where ICL on Qwen achieves the highest similarity, suggesting that Qwen has exceptional paraphrase capabilities when provided in-context demonstrations.

\subsection{Instruction Prompts} \label{apx:prompts}

We include the details of the instructions passed to the various open-weight models during counterfactual generation. The base counterfactual prompt is shown both including the reference summary (Figure~\ref{apx:prompt_ref}) and excluding it (Figure~\ref{fig:exc_prompt}) for two distinct criteria. The Identify-and-Replace (I\&R) prompt is shown in Figure~\ref{fig:ir_prompt}, illustrated on the DREsS ``Language'' criterion. Format is identical for other criteria with rubric details replaced.

\subsection{Tuning Gumbel Noise Control} \label{apx:tuning}

\begin{figure}
        \centering
        \begin{subfigure}{\linewidth}
            \centering
            \includegraphics[width=\linewidth]{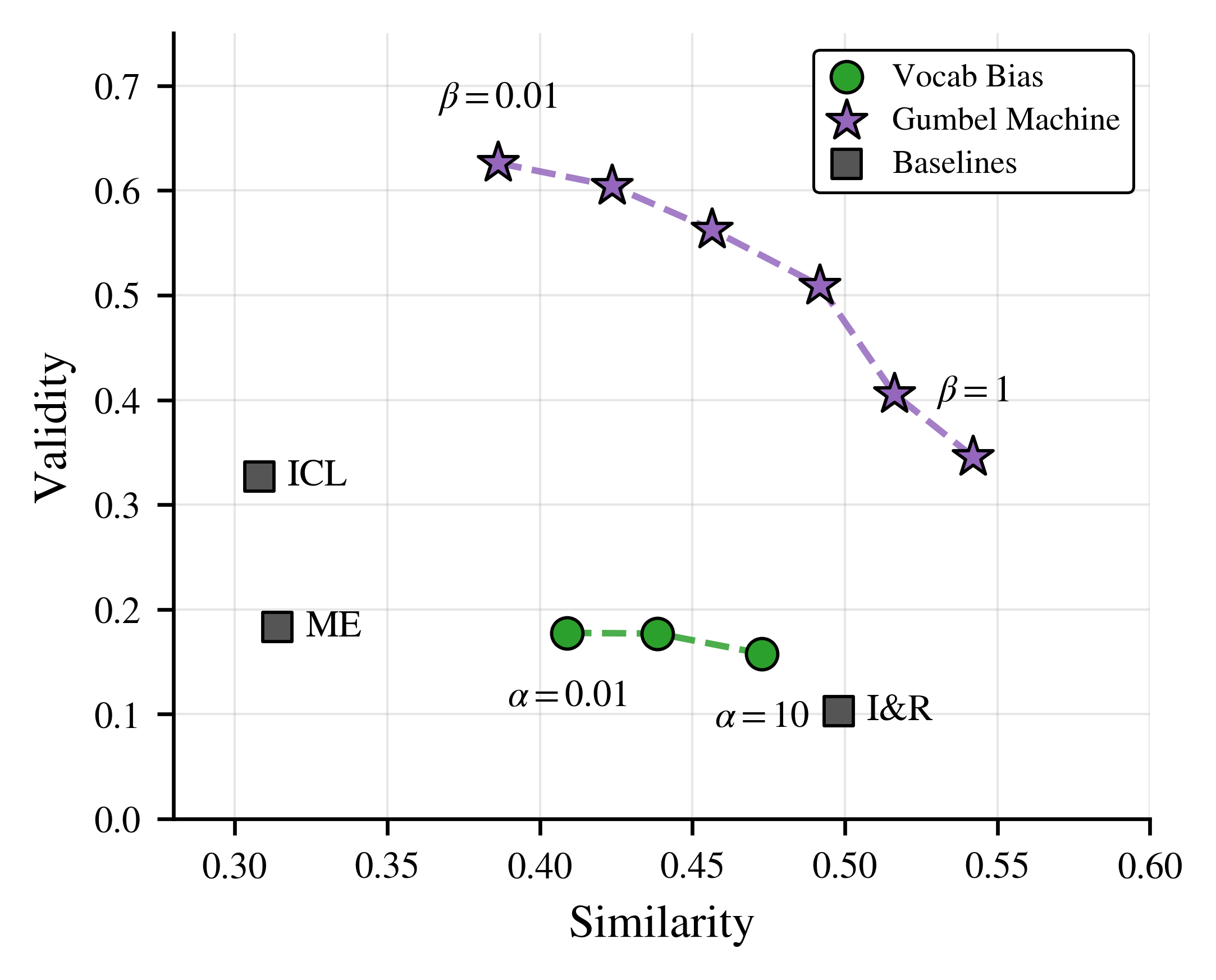}
            \caption{CLASSE}
            \label{fig:pareto-front-classe}
        \end{subfigure}

        \vspace{0.6em}

        \begin{subfigure}{\linewidth}
            \centering
            \includegraphics[width=\linewidth]{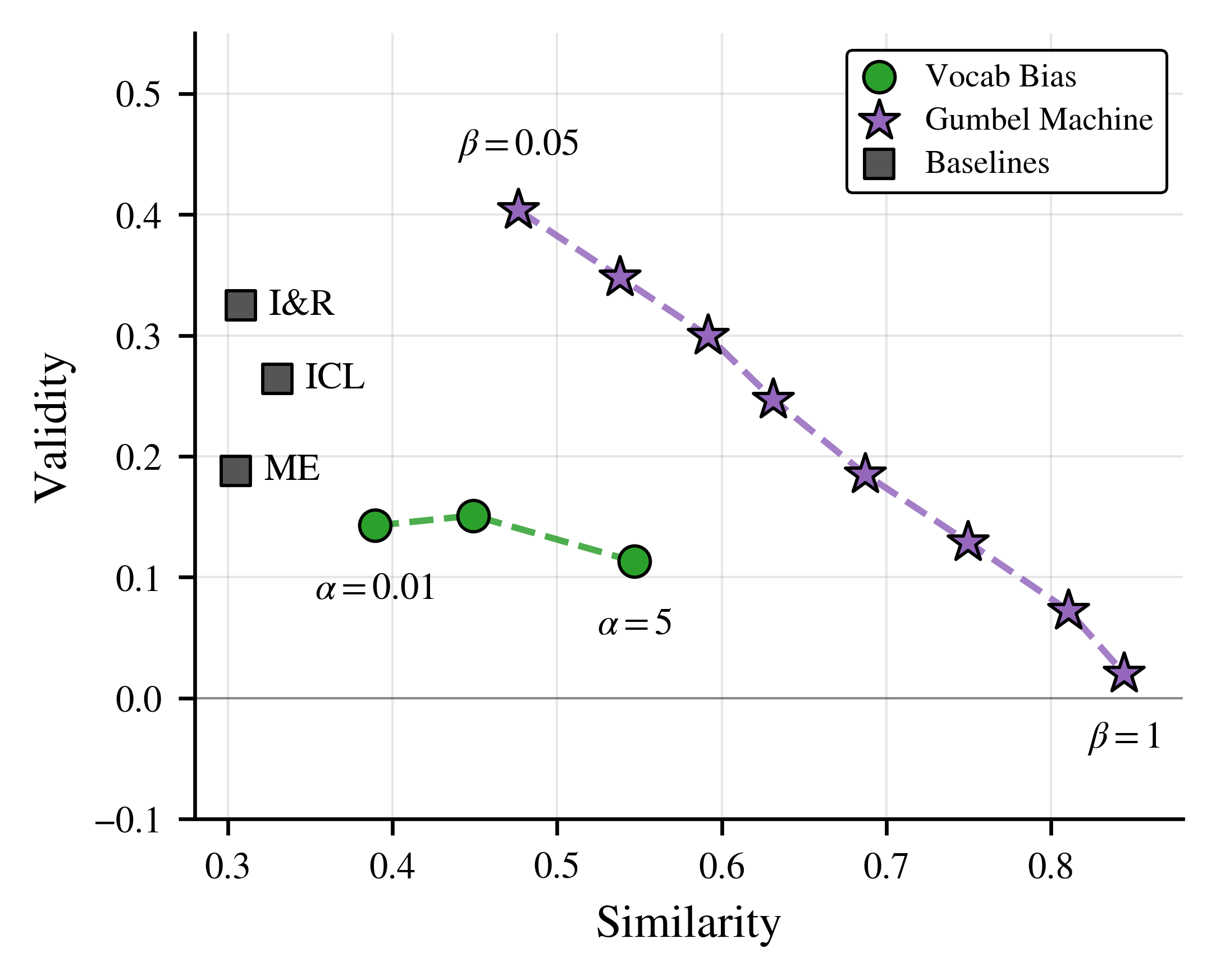}
            \caption{DREsS}
            \label{fig:pareto-front-dress}
        \end{subfigure}
        \caption{Similarity–validity trade-offs comparing Gumbel Machine with a logit bias and baseline approaches (ME - Minimal-Edit, ICL - In-Context Learning, I\&R - Identify-and-Replace) on CLASSE (top) and DREsS (bottom). Gumbel Machine provides a tunable tradeoff between competing metrics, occupying a different region of the Pareto frontier than the logit-bias baseline.}
        \label{fig:pareto-front}
\end{figure}

In Figure~\ref{fig:pareto-front}, we report the results of our experiments to tune the controllable $\beta$ parameter in the GM. In these experiments, we vary the control signal across a range of values for GM and compare with Vocab Bias, a simple decoding strategy that adds a tunable $\alpha$ to all logits for all tokens in the reference summary \cite{yang_fudge_2021}. For GM, we sweep the control parameter $\beta$ across the values $\beta \in \{0.001, 0.1, 0.2, 0.3, 0.4, 1.0\}$ for CLASSE and $\beta \in \{0.05, 0.2, 0.4, 0,5, 0.6, 0.7, 0.8, 1.0\}$ for DREsS. Higher values of $\beta$ emphasize similarity to the original student response, while lower values prioritize satisfying the desired target score. For Vocab Bias, we sweep the control parameter $\alpha$ across the values $\alpha \in \{0.01, 5.0, 10.0\}$ for CLASSE and $\alpha \in \{0.01, 1.0, 5.0\}$ for DREsS. Higher values of $\alpha$ induce stronger similarity to reference tokens. We observe that values outside these ranges result in collapses in model fluency due to excessive logit manipulation.

Comparing the two approaches, we observe that for GM, increasing values of $\beta$ enable a trade-off between similarity that can explore a much wider range of the Pareto front compared to Vocab Bias. Further, we observe that GM tends to maintain higher validity with increasing similarity values. These observations suggest that recovered Gumbel noise provides a more effective control signal for generating similar counterfactual randomness than using token-level heuristics for similarity control. This finding suggests that similarity control in LLMs is more robust when the entire sampling trajectory and random states are considered rather than solely token-level alignment.

\subsection{Scoring Model} \label{apx:scoring}

\begin{table*}[h]
\centering
\small
\setlength{\tabcolsep}{6pt}
\scalebox{.9}{
\begin{tabular}{ll  c  ccc}
\toprule
& & \textbf{BERT} & \multicolumn{3}{c}{\textbf{GPT-4.1 family}} \\
\cmidrule(lr){3-3} \cmidrule(lr){4-6}
\textbf{Dataset} & \textbf{Criterion} & \texttt{ModernBERT}-base & \texttt{GPT-4.1} (prompted) & \texttt{GPT-4.1-mini} (prompted) & \texttt{GPT-4.1-mini} (fine-tuned) \\
\midrule
\multirow{6}{*}{CLASSE}
  & Main Idea     & 0.566 & 0.632 & 0.590 & \textbf{0.683} \\
  & Details       & 0.616 & 0.473 & 0.476 & \textbf{0.690} \\
  & Organization  & 0.607 & 0.495 & 0.393 & \textbf{0.628} \\
  & Wording       & 0.542 & 0.463 & 0.366 & \textbf{0.522} \\
  & Language      & 0.423 & 0.543 & 0.497 & \textbf{0.598} \\
  & \textbf{Mean} & 0.551 & 0.521 & 0.464 & \textbf{0.624} \\
\midrule
\multirow{4}{*}{DREsS}
  & Content      & 0.540 & 0.211 & 0.289 & \textbf{0.652} \\
  & Organization & 0.490 & 0.278 & 0.213 & \textbf{0.598} \\
  & Language     & 0.475 & 0.208 & 0.252 & \textbf{0.553} \\
  & \textbf{Mean} & 0.502 & 0.232 & 0.251 & \textbf{0.601} \\
\bottomrule
\end{tabular}}
\caption{Per-criterion QWK agreement between scoring models and human raters on the CLASSE and DREsS validation sets.}
\label{apx:scoring_per_criterion}
\end{table*}

Figure~\ref{fig:score_prompt} shows the instructions passed to our scoring model for the ``Language'' criterion. Rubric details are modified relative to each criterion. The model is prompted with top $p=1.0$ and temperature $0$ to reduce randomness, with an instruction to provide justification before scoring to encourage Chain-of-Thought reasoning.

\paragraph{Scoring Model Comparison} Our final scoring model is a fine-tuned \texttt{GPT-4.1-mini} model, trained per criterion. We report scoring-model comparisons across four configurations: (i) a fine-tuned BERT encoder following \cite{fernandez_automated_2022}, (ii) prompting the larger \texttt{GPT-4.1} model, (iii) prompting \texttt{GPT-4.1-mini}, and (iv) fine-tuned \texttt{GPT-4.1-mini}. Table~\ref{apx:scoring_per_criterion} reports per-criterion QWK against human ratings on a held-out validation set for CLASSE and DREsS, respectively. Across both datasets the fine-tuned \texttt{GPT-4.1-mini} achieves the highest mean QWK (CLASSE $0.624$, DREsS $0.601$) and outperforms every other scoring configuration on nearly every criterion.

For the BERT scoring model, we fine-tune a \texttt{ModernBERT-base} encoder with a linear regression head over the integer rubric scores using mean-squared-error loss. For both prompted and fine-tuned GPT models we use the template shown in Figure~\ref{fig:score_prompt}, sampled with temperature $0$ and top-$p=1.0$. For fine-tuning, we use OpenAI's hosted fine-tuning API on \texttt{gpt-4.1-mini-2025-04-14}, training one checkpoint per criterion on the criterion-specific train split with the same prompt template used at inference.

\subsection{Prompt Ablation} \label{apx:prompt-ablate}

\begin{table}[t]
\centering
\small
\begin{tabular}{lcc}
\toprule
\textbf{Criterion} & \textbf{Reference} & \textbf{No Reference} \\
\midrule
Main Idea  & 0.626 & 0.347 \\
Details    & 0.604 & 0.315 \\
Cohesion   & 0.597 & 0.315 \\
Wording    & 0.611 & 0.352 \\
Language   & 0.622 & 0.325 \\
\midrule
\textbf{Overall} & 0.612 & 0.331 \\
\bottomrule
\end{tabular}
\caption{Ablation study on reference conditioning influence on recovered Gumbel noise control effectiveness. Table reports average similarity scores across counterfactuals. Across all criteria, reference hinting shows higher effectiveness of the recovered noise.}
\label{tab:foresight_ablation}
\end{table}

In Table~\ref{tab:foresight_ablation}, we report results of an ablation study where we compare how inclusion and exclusion of the reference summary in the prompt influence the effectiveness of $\beta$-Hindsight control. In this ablation, we include/exclude the reference both during noise recovery and replay, set $\beta=1.0$, and run hindsight sampling on the base Llama-3 setting. We see an average counterfactual similarity score of $0.331$ when excluding the reference, which is similar to ME. However, when including the reference at recovery, the generated counterfactual summary is much more similar to the reference. These results suggest a nuanced interaction between the replayed Gumbel noise and the conditioning context, highlighting the importance of reference inclusion for effective $\beta$-Hindsight control. One possible explanation is that, when the reference tokens are highly certain (logit value of the reference token significantly exceeds that of other ones), the recovered noise is not strong enough to significantly alter the LLM's overall output trajectory, and only provides small adjustments at a given timestep. On the contrary, when the reference tokens are less certain, the recovered noise encodes nuanced information, resulting in a volatile control signal with conflicting objectives that can significantly alter the output trajectory. 

\subsection{Dataset Details} \label{apx:classe}

\begin{figure*}
    \centering
    \includegraphics[width=1\linewidth]{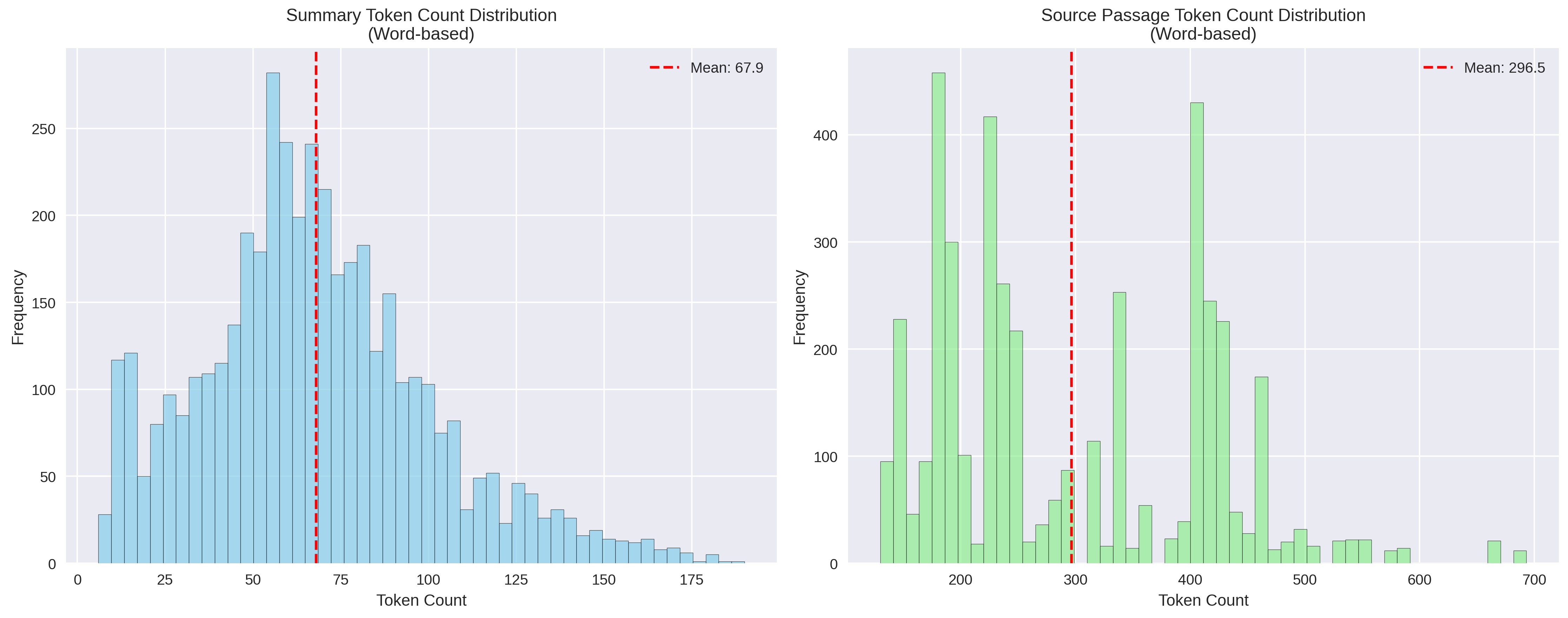}
    \caption{Token count statistics for the summaries and passages in the CLASSE dataset. Summaries are approximately 68 words in length and passages are approximately 297 words.}
    \label{fig:token_counts}
\end{figure*}

\begin{figure*}
    \centering
    \includegraphics[width=1\linewidth]{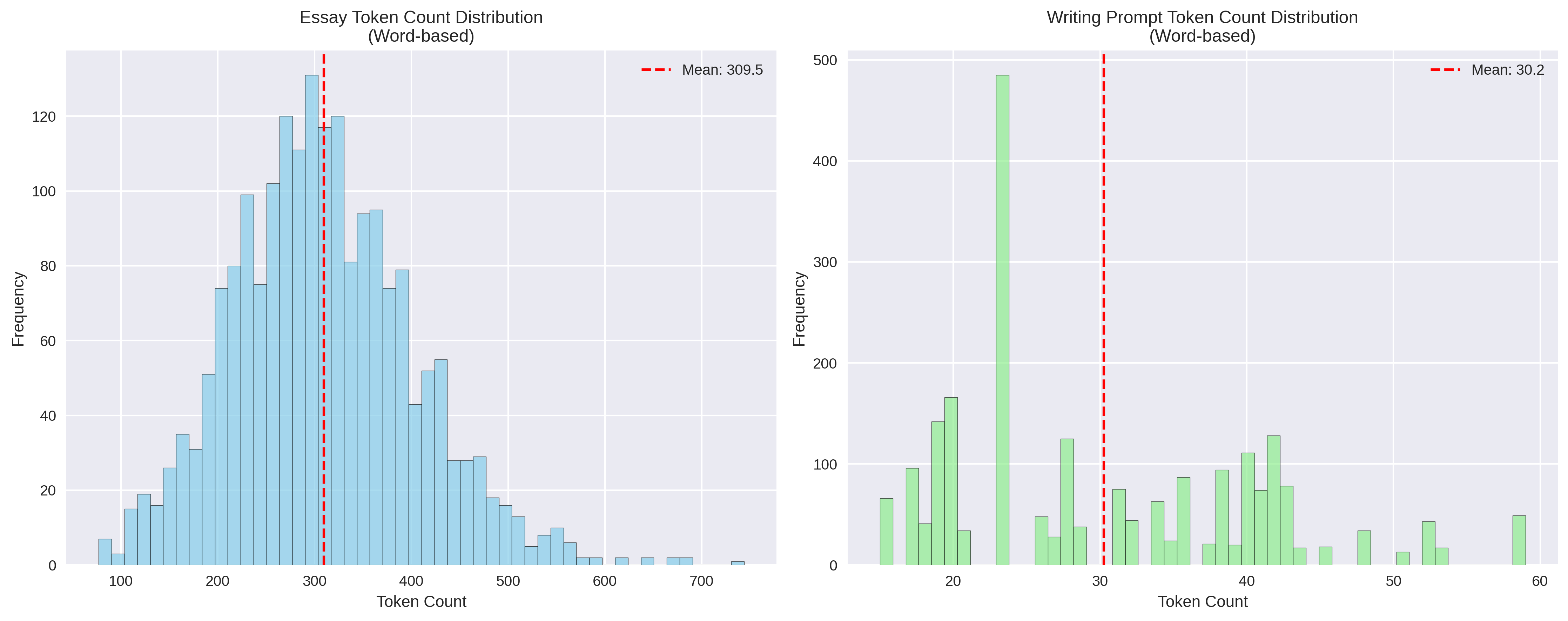}
    \caption{Token count statistics for the essays and writing prompts in the DREsS dataset. Essays are approximately 310 words in length and writing prompts are approximately 30 words.}
    \label{fig:dress_token_counts}
\end{figure*}

We report detailed descriptions for each CLASSE criterion in Table~\ref{apx:criteria} and Table~\ref{apx:rubric} details score definitions for each possible value. Figure~\ref{fig:token_counts} provides a breakdown of the summary and passage lengths in the dataset. Summaries are approximately 68 words in length and passages are approximately 297 words (calculated by simple space separation). We report detailed descriptions for each DREsS criterion in Table~\ref{apx:dress_criteria} and Table~\ref{apx:dress_rubric} details score definitions for each possible value. Figure~\ref{fig:dress_token_counts} provides a breakdown of the essay and writing-prompt lengths in the dataset. Essays are approximately 310 words in length and writing prompts are approximately 30 words.

\subsection{Training Details} \label{apx:training}

We train all GM checkpoints using parameter-efficient fine-tuning on \texttt{LLaMA-3.1-8B-Instruct} and \texttt{Qwen3-4B-Instruct-2507}. The pipeline runs supervised fine-tuning (SFT) on counterfactual prompt-target pairs, followed by direct preference optimization (DPO) on score-gap preference pairs constructed from the same training split. For both stages, we train a rank-$32$ LoRA adapter with dropout $0.05$ for all attention and MLP projections using an AdamW optimizer. For SFT, we use a constant learning rate of $5\!\times\!10^{-5}$ with weight decay $0.01$, and in DPO we use a cosine schedule at $2\!\times\!10^{-5}$ against the frozen SFT LoRA as the reference policy.

We estimate our experiments to spend 350 NVIDIA GPU-hours across A100, L40, A40, and RTX 8000 GPUs. In addition to closed-weight API access to Claude Haiku 4.5, Claude Sonnet 4.6, and 8 fine-tuned GPT-4.1-mini judges.


\subsection{Human Evaluation Details} \label{sec:apdx-human-eval}


\paragraph{Task structure.} Each evaluator was shown 50 (passage, student summary, rewrite A, rewrite B) tuples on the CLASSE dataset, stratified across the five criteria and across the original student score on the target criterion. The pair of rewrites was sampled one from GM and one from the Minimal-Edit baseline at matched automated validity (target score $4$); ordering was randomized and the generating approach was hidden from the evaluator. For each tuple, evaluators answered three per-rewrite questions independently for A and B on a 1--3 scale (None / Slight / Clear), and then three direct-comparison questions with options $\{A, B, \text{Same}\}$. The task took evaluators roughly 2 hours, and we paid them \$60 each, which is significantly higher than the minimum wage in the country where this study was conducted. All evaluators gave consent before participating via an IRB-approved consent form.

\paragraph{Calibration material.} Evaluators were provided with a reference document, available throughout the task, that contained: (i) a description of the rewrite-comparison task and a worked positive/negative rewrite example; (ii) detailed guidance on each of the six questions with concrete anchors for the 1--3 scale; (iii) a sample passage paired with each rubric criterion; and (iv) for each criterion, the score-level rubric (1--4) together with an example student summary at every rubric score for calibration. Table~\ref{apx:rubric} and Table~\ref{apx:criteria} provide detailed definitions shown to evaluators for each criterion.

\begin{figure*}[h]
    \centering
    \includegraphics[width=1\linewidth]{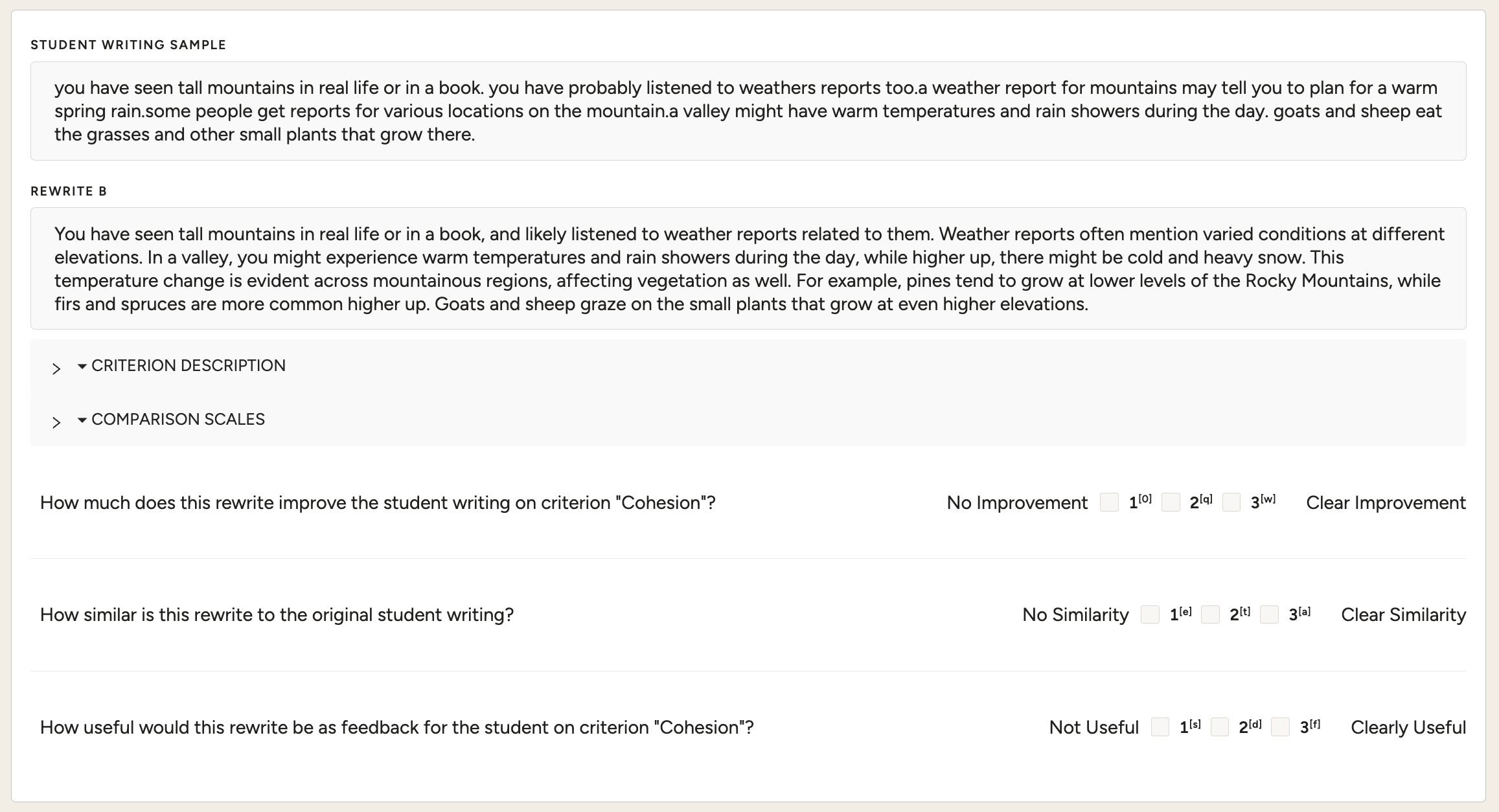}
    \caption{Teacher evaluation interface per-rewrite questions (Q1--Q3) for rewrites A and B.}
    \label{fig:label_interface_1}
\end{figure*}

\begin{figure*}[h]
    \centering
    \includegraphics[width=1\linewidth]{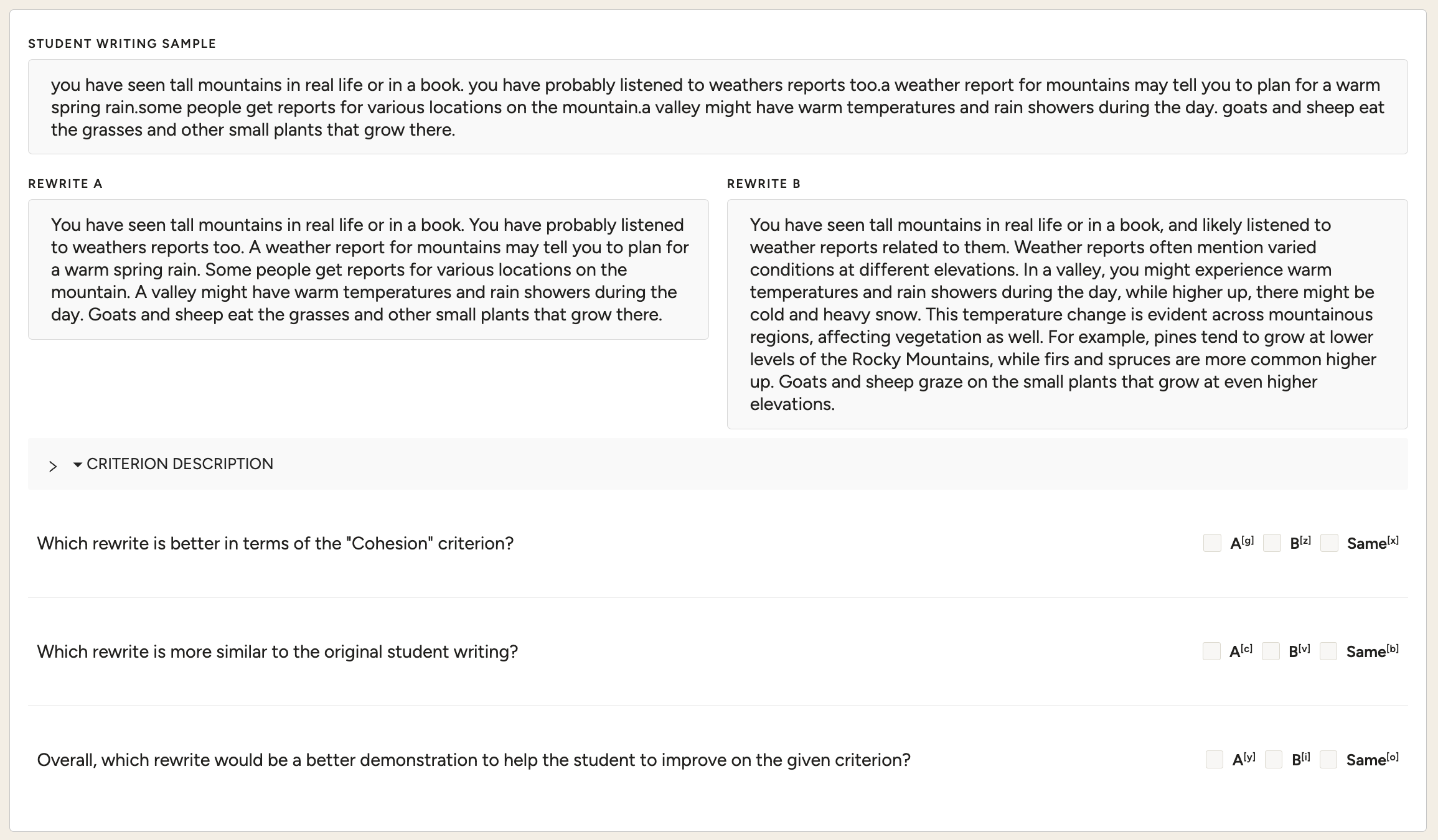}
    \caption{Teacher evaluation interface pairwise comparison questions (Q4--Q6).}
    \label{fig:label_interface_2}
\end{figure*}

\paragraph{Interface.} Figures~\ref{fig:label_interface_1} and~\ref{fig:label_interface_2} show the evaluation interface used in the study. The interface presents the passage and the student summary alongside the two rewrites, with the per-rewrite questions appearing first (Figure~\ref{fig:label_interface_1}) and the direct-comparison questions appearing second (Figure~\ref{fig:label_interface_2}).

\paragraph{Inter-rater Agreement} We observe an inter-rater agreement (Krippendorff's $\alpha$) on the similarity rating Likert of $\alpha=0.29$ across all four raters, increasing to $\alpha=0.43$ when restricted to the three raters who actively differentiated on this dimension (one rater frequently selected "Same" on similarity comparisons, 36/50 tasks). The directional finding that GM produces more similar rewrites is replicated by all four raters individually.

\onecolumn
\subsection{Qualitative Counterfactual Examples} \label{apx:qualitative}

\input{examples/civil_service}

\clearpage

\input{examples/global_warming}

\clearpage

\input{examples/ecological_pyramids}
\twocolumn

\input{algo_v2}


\input{prompts/language_ref}

\input{prompts/details_excl_ref}

\input{prompts/ir_dress}

\input{prompts/scoring}


\begin{table*}[!h]
\centering
\renewcommand{\arraystretch}{1.4}
\begin{tabular}{|p{3.5cm}|p{10cm}|}
\hline
\textbf{Criterion} & \textbf{Description} \\
\hline
Main Idea &
This criterion measures how well the main idea of the summary relates to the central topic of the passage. In an ideal summary, the two are closely aligned, and a topic sentence appears at the beginning that clearly captures the main idea. \\
\hline
Details &
This criterion measures how thoroughly and precisely the summary covers the key information discussed in the passage. In an ideal summary, all key information is included, with no irrelevant content or inaccurate representation of the ideas in the passage. \\
\hline
Organization &
This criterion measures how well organized and naturally constructed the summary is. In an ideal summary, each sentence flows naturally into the subsequent sentence without an unnatural or abrupt shift in idea or topic. \\
\hline
Wording &
This criterion measures how well the summary reflects original wording from the passage. While some verbatim copying may be unavoidable for accuracy, an ideal summary paraphrases the passage with minimal direct copying. \\
\hline
Language &
This criterion measures how well the original aspects of the summary demonstrate mastery of English grammar and vocabulary. Portions of the summary that are not verbatim copies should be assessed. Writers range from grades 3 to 12; therefore, sophisticated adult-level language is not expected. \\
\hline
\end{tabular}
\caption{Summary Evaluation Criteria Descriptions}
\label{apx:criteria}
\end{table*}

\begin{table*}[h]
\centering
\renewcommand{\arraystretch}{1.4}
\begin{tabular}{|p{3.5cm}|p{2cm}|p{8.5cm}|}
\hline
\textbf{Criterion} & \textbf{Score} & \textbf{Descriptor} \\
\hline
\multirow{4}{=}{Main Idea}
& 1 (Poor) & Main idea is not linked to the central topic. \\
& 2 (Fair) & Main idea is linked to the central topic, but there is no topic sentence to bring ideas together. \\
& 3 (Good) & Main idea is linked to the central topic and includes a topic sentence stating some aspect of the content. \\
& 4 (Excellent) & Main idea is linked to the central topic and includes a topic sentence that clearly states the main idea. \\
\hline
\multirow{4}{=}{Details}
& 1 (Poor) & Statements are not related to the passage. \\
& 2 (Fair) & Some key information from the passage is included, but important ideas are missing. \\
& 3 (Good) & Most key information is included, but some ideas may be irrelevant or inaccurate. \\
& 4 (Excellent) & All key information from the passage is included with no irrelevant ideas. \\
\hline
\multirow{4}{=}{Organization}
& 1 (Poor) & Ideas are randomly presented and do not link to each other. \\
& 2 (Fair) & Some ideas link to each other. \\
& 3 (Good) & Most ideas are logically presented. \\
& 4 (Excellent) & All ideas are logically presented. \\
\hline
\multirow{4}{=}{Wording}
& 1 (Poor) & Summary shows a heavy reliance on verbatim copying of source language. \\
& 2 (Fair) & Summary shows some use of original wording, but includes verbatim or near-copying of source language. \\
& 3 (Good) & Summary shows evidence of appropriate levels of paraphrasing. \\
& 4 (Excellent) & Summary shows substantial evidence of appropriate paraphrasing use. \\
\hline
\multirow{4}{=}{Language}
& 1 (Poor) & Summary shows a very basic understanding of lexical and syntactic structures. \\
& 2 (Fair) & Summary shows an understanding of lexical and syntactic structures. \\
& 3 (Good) & Summary shows an appropriate range of lexical and syntactic structures. \\
& 4 (Excellent) & Summary shows an excellent range of lexical and syntactic structures. \\
\hline
\end{tabular}
\caption{Detailed definitions and rubrics for each criterion in the CLASSE dataset}
\label{apx:rubric}
\end{table*}

\begin{table*}[!h]
\centering
\renewcommand{\arraystretch}{1.4}
\begin{tabular}{|p{3.5cm}|p{10cm}|}
\hline
\textbf{Criterion} & \textbf{Description} \\
\hline
Content &
This criterion measures how well the essay develops and supports its argument with relevant reasoning and examples. An ideal essay presents well-developed ideas that clearly and thoroughly support the argument. \\
\hline
Organization &
This criterion measures how clearly and logically the essay is structured. An ideal essay has a clear, easy-to-follow argument with effective transitions and coherence devices throughout. \\
\hline
Language &
This criterion measures the student's control of vocabulary and grammatical structures in the essay. An ideal essay demonstrates a wide, accurate range of vocabulary and consistently correct grammar, spelling, and punctuation. \\
\hline
\end{tabular}
\caption{Essay Evaluation Criteria Descriptions for the DREsS dataset.}
\label{apx:dress_criteria}
\end{table*}

\begin{table*}[h]
\centering
\renewcommand{\arraystretch}{1.4}
\begin{tabular}{|p{3.5cm}|p{2.4cm}|p{8.1cm}|}
\hline
\textbf{Criterion} & \textbf{Score} & \textbf{Descriptor} \\
\hline
\multirow{5}{=}{Content}
& 1 (Very Poor) & Paragraph lacks relevance to the argument; little to no supporting reasoning or examples. \\
& 2 (Poor) & Minimal development; ideas unclear or loosely related to argument; weak or vague examples. \\
& 3 (Fair) & Addresses the argument but has limited or uneven development; general reasons with insufficient examples. \\
& 4 (Good) & Relevant and mostly well-developed; clear reasons and some effective examples. \\
& 5 (Excellent) & Well-developed and relevant; strong reasons and examples that clearly support the argument. \\
\hline
\multirow{5}{=}{Organization}
& 1 (Very Poor) & No clear structure; disjointed or random ideas; very difficult to follow. \\
& 2 (Poor) & Difficult to follow; poorly sequenced; limited or ineffective transitions; unclear focus. \\
& 3 (Fair) & Apparent but inconsistent organization; loosely connected ideas; unclear focus or transitions. \\
& 4 (Good) & Clear and logical structure; effective transitions and coherence devices; minor lapses in flow or focus. \\
& 5 (Excellent) & Very effectively structured; easy to follow ideas and argument building; effective coherence devices throughout. \\
\hline
\multirow{5}{=}{Language}
& 1 (Very Poor) & Very basic or incorrect language; pervasive errors severely hinder understanding. \\
& 2 (Poor) & Limited vocabulary; frequent grammar, spelling, or punctuation errors affecting readability. \\
& 3 (Fair) & Sufficient but limited vocabulary; noticeable errors occasionally affect clarity. \\
& 4 (Good) & Wide range of vocabulary with generally accurate usage; minor errors do not affect clarity. \\
& 5 (Excellent) & Sophisticated control of wide vocabulary and collocations; correct grammar, spelling, and punctuation throughout. \\
\hline
\end{tabular}
\caption{Detailed definitions and rubrics for each criterion in the DREsS dataset.}
\label{apx:dress_rubric}
\end{table*}


\end{document}

%% file: algo_v2_short.tex
\begin{algorithm}[t]
\caption{$\beta$-Hindsight Control}
\label{alg:gumbel-cf-short}
\begin{algorithmic}[1]

\STATE \textbf{def} BetaHindsight$(\mathbf x, \mathbf y, \mathbf x', \theta, \beta)$:
\STATE \quad $\mathbf G \gets \varnothing$

\STATE \quad \textbf{for} $t = 1,\dots,|\mathbf y|$: \algcmt{Recover Noise}
\STATE \qquad $\ell \gets \mathrm{LLM}(\mathbf x, \mathbf y_{<t};\theta)$
\STATE \qquad $\ell_{max} \gets \max_{v\in V} \ell[v]$
\STATE \qquad $\mathbf g_t[y_t] \sim \mathrm{LowerTruncG}(\ell_{max} - \ell[y_t])$
\STATE \qquad $T_t \gets \ell[y_t] + \mathbf g_t[y_t]$
\STATE \qquad \textbf{for} $v \in V \setminus \{y_t\}$:
\STATE \qquad\quad $\mathbf g_t[v] \sim \mathrm{UpperTruncG}(T_t - \ell[v])$
\STATE \qquad store $\mathbf g_t$ in $\mathbf G$

\STATE \quad $\mathbf y' \gets \varnothing$
\STATE \quad $y'_1 \gets \mathrm{BOS},\; t \gets 1$

\STATE \quad \textbf{while} $y'_t \neq \mathrm{EOS}$: \algcmt{Replay noise}
\STATE \qquad $\ell \gets \mathrm{LLM}(\mathbf x', \mathbf y'_{<t}; \theta)$
\STATE \qquad \textbf{if} $t \le |\mathbf y|$:
\STATE \qquad\quad $\mathbf g_t \gets \mathbf G[t]$
\STATE \qquad\quad $y'_t \gets \arg\max_{v\in V}\bigl(\ell[v] + \beta \cdot \mathbf g_t[v]\bigr)$
\STATE \qquad \textbf{else}:
\STATE \qquad\quad $\mathbf u_t \sim \mathrm{Gumbel}(0,1)^{|V|}$
\STATE \qquad\quad $y'_t \gets \arg\max_{v\in V}\bigl(\ell[v] + \mathbf u_t[v]\bigr)$
\STATE \qquad append $y'_t$ to $\mathbf y'$, $t \gets t+1$

\STATE \quad \textbf{return} $\mathbf y'$

\end{algorithmic}
\end{algorithm}

%% file: examples/civil_service.tex
\begin{examplebox}[label=ex:civil-service]{Example 1 --- Civil Service}
\textbf{Student writing} (Score 3):

\begin{quote}\itshape
Civil service jobs are available for all people.\,\, These jobs have fair pay from the merit system.\,\, These jobs are filled by the top applicants.\,\, This process is time consuming, but he benefits to having a civil service job out ways the steps that need to be taken.
\end{quote}

\textbf{GM rewrite} (sim $0.50$):

\begin{quote}\itshape
Civil service jobs are available for all people, offering a wide range of positions from clerks to law enforcement. These jobs are filled by the top applicants through the merit system, which ensures fair pay and fair treatment for all employees. The merit system is set up to protect the employee and provide equal and fair pay. This process is time consuming, but the benefits of having a civil service job outweigh the steps that need to be taken.
\end{quote}

\textbf{ME rewrite} (sim $0.30$):

\begin{quote}\itshape
Civil service jobs offer opportunities to men and women at various levels of government, including city, county, state, and federal. The merit system is used to recruit the most qualified individuals for each position, ensuring fair pay, training, and treatment regardless of background. With a range of job types available, from administrative roles to law enforcement and skilled trades, civil service offers secure employment options for those willing to invest the time and effort required to secure a position.
\end{quote}
\end{examplebox}
\captionof{figure}{Example of a counterfactual rewrite where Gumbel Machine (GM) remains similar in precise wording and structure to the student writing while inserting relevant intermediate sentences and clauses to improve the summary. Minimal Edit (ME) instead reorganizes the content significantly and does not preserve the student's original wording.}
\label{ex:civil-service}

%% file: examples/global_warming.tex
\begin{examplebox}{Example 2 --- Global Warming}
\textbf{Student writing} (Score 1):

\begin{quote}\itshape
Most people in California are more environmentally conscious than in other places in the country. They also don't believe that the government is doing enough to stop this.
\end{quote}

\textbf{GM rewrite} (sim $0.24$):

\begin{quote}\itshape
Most Californians are very concerned about global warming. They believe that the earth is warming too much and that it will affect living things on the earth. A survey was taken in California to see how people felt about global warming. The results showed that most people in California are very concerned about global warming and want the state to work on the problem. They are not happy with the federal government's handling of the issue and want the state to start working on it immediately.
\end{quote}

\textbf{ME rewrite} (sim $0.19$):

\begin{quote}\itshape
California residents are particularly concerned about the state of the planet, linking the issue to global warming. They believe that this warming poses a threat to living things. Californians want the state to take responsibility for investigating the problem.

Californians are deeply disturbed when it comes to the federal government's response to global warming. This year, they are more worried than ever, feeling the problem is getting worse. They want the state to take action immediately.

Californians want the state to start working on the problem as soon as possible, rather than waiting for the federal government to act.
\end{quote}
\end{examplebox}
\captionof{figure}{Example of a counterfactual rewrite where Gumbel Machine (GM) preserves the student's two-part key points, but then expands on the writing to provide more details. Minimal Edit (ME) instead fragments this structure across three paragraphs and introduces new temporal framing absent from the original.}
\label{ex:global-warming}

%% file: examples/ecological_pyramids.tex
\begin{examplebox}{Example 3 --- Ecological Pyramids}
\textbf{Student writing} (Score 3):

\begin{quote}\itshape
Ecologists construct a pyramid of numbers based on the population sizes of organisms in each trophic level. Sometimes population sizes decrease at each higher trophic level. This is not always true. For example, one tree can be food for 50,000 insects. In this case, the pyramid would be inverted.

Energy decreases at each succeeding trophic level. The total energy transfer from one trophic level to the next is only about ten percent because organisms fail to capture and eat all the food available at the trophic level below them. When an organism consumes food, it uses some of the energy in the food for metabolism, some for building body tissues, and some is given off as waste.
\end{quote}

\textbf{GM rewrite} (sim $0.75$):

\begin{quote}\itshape
Ecologists construct a pyramid of numbers based on the population sizes of organisms in each trophic level. Sometimes population sizes decrease at each higher trophic level. This is not always true. For example, one tree can be food for 50,000 insects. In this case, the pyramid would be inverted. Energy decreases at each succeeding trophic level. The total energy transfer from one trophic level to the next is only about ten percent because organisms fail to capture and eat all the food available at the trophic level below them. When an organism consumes food, it uses some of the energy in the food for metabolism, some for building body tissues, and some is given off as waste. When the organism is eaten, the energy that was used to build body tissue is available as energy to be used by the organism that consumed it. The energy lost at each successive trophic level enters the environment as heat.
\end{quote}

\textbf{ME rewrite} (sim $0.34$):

\begin{quote}\itshape
Ecologists use ecological pyramids to model the distribution of matter and energy within an ecosystem. This type of pyramid shows how energy flows through an ecosystem. The initial source of energy for all ecological pyramids is energy from the sun, which decreases at each succeeding trophic level due to inefficient energy transfer. The total energy transfer from one trophic level to the next is only about ten percent. When an organism consumes food, it uses some of the energy in the food for metabolism, some for building body tissues, and some is given off as waste, with the energy lost at each successive trophic level entering the environment as heat. Ecologists also construct pyramids of numbers based on population sizes, which sometimes decrease at each higher trophic level, but can also be inverted, such as when one tree is food for 50,000 insects.
\end{quote}
\end{examplebox}
\captionof{figure}{Example of a counterfactual rewrite where Gumbel Machine (GM) retains the student's original wording and paragraph structure almost entirely, appending only a small number of sentences to improve the response. Minimal Edit (ME) instead reorganizes the content by foregrounding a general definition, reordering topics, and introducing new framing devices not present in the student writing.}
\label{ex:ecological-pyramids}

%% file: algo_v2.tex
\begin{algorithm*}
\caption{Expanded $\beta$-Hindsight Control Algorithm with Comments}
\label{alg:gumbel-cf-long}
\begin{algorithmic}[1]

\STATE \textbf{def} RecoverNoise$(\mathbf x, \mathbf y, \theta)$:
\STATE \quad $\mathbf G \gets \varnothing$
\STATE \quad \textbf{for} $t = 1,\dots,|\mathbf y|$:
\STATE \qquad $\ell \gets \mathrm{LLM}(\mathbf x, \mathbf y_{<t};\theta)$ \algcmt{compute logits}
\STATE \qquad $\ell_{max} \gets \max_{v\in V} \ell[v]$ 
\STATE \qquad $\mathbf g_t[y_t] \sim \mathrm{LowerTruncG}(\ell_{max} - \ell[y_t])$
        \algcmt{recover noise for observed token}
\STATE \qquad $T_t \gets \ell[y_t] + \mathbf g_t[y_t]$ 
\STATE \qquad \textbf{for} $v \in V \setminus \{y_t\}$:
\STATE \qquad\quad $\mathbf g_t[v] \sim \mathrm{UpperTruncG}(T_t - \ell[v])$
        \algcmt{recover shifted noise}
\STATE \qquad store $\mathbf g_t$ in $\mathbf G$
\STATE \quad \textbf{return} $\mathbf G$
\vspace{6pt}

\STATE \textbf{def} GenerateCF$(\mathbf x', \mathbf G, \mathbf y, \theta, \beta)$:
\STATE \quad $\mathbf y' \gets \varnothing$
\STATE \quad $y'_1 \gets \mathrm{BOS},\; t \gets 1$
\STATE \quad \textbf{while} $y'_t \neq \mathrm{EOS}$:
\STATE \qquad $\ell \gets \mathrm{LLM}(\mathbf x', \mathbf y'_{<t}; \theta)$
        \algcmt{logits under intervened model}
\STATE \qquad \textbf{if} $t \le |\mathbf y|$:
\STATE \qquad\quad $\mathbf g_t \gets \mathbf G[t]$ \algcmt{reuse recovered noise}
\STATE \qquad\quad $y'_t \gets \arg\max_{v\in V}\bigl(\ell[v] + \beta \cdot \mathbf g_t[v]\bigr)$
        \algcmt{$\beta$-scaled noise Gumbel-max sample}
\STATE \qquad \textbf{else}:
\STATE \qquad\quad $\mathbf u_t \sim \mathrm{Gumbel}(0,1)^{|V|}$
        \algcmt{standard Gumbel-max sample beyond original length}
\STATE \qquad\quad $y'_t \gets \arg\max_{v\in V}\bigl(\ell[v] + \mathbf u_t[v]\bigr)$
\STATE \qquad append $y'_t$ to $\mathbf y'$, $t \gets t+1$
\STATE \quad \textbf{return} $\mathbf y'$
\vspace{6pt}

\STATE \textbf{def} BetaHindsight$(\mathbf x, \mathbf y, \mathbf x', \theta, \beta)$:
\STATE \quad $\mathbf G \gets$ RecoverNoise$(\mathbf x, \mathbf y, \theta)$
        \algcmt{recover posterior noise}
\STATE \quad $\mathbf y' \gets$ GenerateCF$(\mathbf x', \mathbf G, \mathbf y, \theta, \beta)$
        \algcmt{construct counterfactual via replay}
\STATE \quad \textbf{return} $\mathbf y'$

\end{algorithmic}
\end{algorithm*}

%% file: prompts/language_ref.tex
\begin{figure*}
\begin{promptbox}{Example 1: ``Language'' Counterfactual Instruction Prompt}
Your task is to summarize a passage in a way which would receive a score following a set criteria.

You will be given the rubric, the desired scores for each part of the rubric, a passage to summarize, and a reference summary which received a different score.
Use this information to write a summary which would correspond to the given scores and while keeping your summary as close to the reference as possible.

\#\#\# Rubric:

\#\#\# Language Beyond Source Text

**Scores:**

- 1: Poor

- 2: Fair

- 3: Good

- 4: Excellent

**Score Descriptions:**

- **1**: Summary shows a very basic understanding of lexical and syntactic structures.

- **2**: Summary shows an understanding of lexical and syntactic structures.

- **3**: Summary shows an appropriate range of lexical and syntactic structures.

- **4**: Summary shows an excellent range of lexical and syntactic structures.

\#\#\# Passage:

\{\{ reading\_passage\_text \}\}

\#\#\# Desired scores:

**Language Beyond Source Text:** \{\{ language\_score \}\}

\#\#\# Reference Summary:

\{\{ original\_student\_summary \}\}

End your response directly after giving the summary

Do not provide explanation, justification, or notes before or after the summary.

Do not preface with "Here is the summary..."

Do not postface the summary with ``Note'':

\#\#\# Summary:
\end{promptbox}
\caption{Example instruction prompt for ``Language'' criterion including reference. This prompt structure is the basis for Gumbel Machine (GM) and Minimal Edit (ME) structures. In-context Learning (ICL) additionally includes examples after the score descriptions (not shown).}
\label{apx:prompt_ref}
\end{figure*}

%% file: prompts/details_excl_ref.tex
\begin{figure*}
\begin{promptbox}{Example 2: ``Details'' Counterfactual Instruction - Excluding Reference}
Your task is to summarize a passage in a way which would receive a score following a set criteria.

You will be given the rubric, the desired score, and a passage to summarize. Use this information to write a summary which would correspond to the given score.

\#\#\# Rubric:

\#\#\# Details

**Scores:**

- 1: Poor

- 2: Fair

- 3: Good

- 4: Excellent

**Score Descriptions:**

- **1**: Statements are not related to the passage.

- **2**: Some key information from the passage is included, but important ideas are missing.

- **3**: Most key information from the passage is included, but some ideas may be irrelevant or inaccurate.

- **4**: All key information in the passage is included without irrelevant ideas.

\#\#\# Passage:

\{\{ reading\_passage\_text \}\}

\#\#\# Desired scores:

**Details:** \{\{ details\_score \}\}

End your response directly after giving the summary

Do not provide explanation, justification, or notes before or after the summary.

Do not preface with "Here is the summary..."

Do not postface the summary with ``Note'':

\#\#\# Summary:
\end{promptbox}
\caption{Example instruction prompt for ``Details'' criterion excluding reference. This prompt structure is used for model training and ablation studies.}
\label{fig:exc_prompt}
\end{figure*}

%% file: prompts/ir_dress.tex
\begin{figure*}
\begin{promptbox}{Example 3: ``Language'' Identify-and-Replace (I\&R) Prompt --- DREsS}
Your task is to write an essay for a given prompt.

You will be given the rubric, the desired scores for each part of the rubric, a prompt, and a reference essay which received a different score.
Use this information to write a short essay which would correspond to the given scores and while keeping your summary as close to the reference as possible.
Your essay should be similar length to the provided reference text.

\#\# Scoring Rubric

**Language Score** (1--5):

- 1: Very basic or incorrect language use; pervasive errors severely hinder understanding.

- 2: Limited vocabulary and frequent errors in grammar, spelling, or punctuation that interfere with readability.

- 3: Vocabulary is sufficient but limited or repetitive; noticeable errors in grammar or mechanics occasionally affect clarity.

- 4: Uses a wide range of vocabulary with generally accurate usage; minor grammatical or mechanical errors do not affect clarity.

- 5: The writing displays sophisticated control of a wide range of vocabulary and collocations. The essay follows grammar and usage rules throughout the essay. Spelling and punctuation are correct throughout the essay.

\#\# Target Essay Prompt

\{\{ prompt \}\}

\#\# Reference Essay

\{\{ gt\_reference \}\}

\#\# Original Scores

**Language Score:** \{\{ original\_language\_score \}\}

\#\# Desired Scores

**Language Score:** \{\{ language\_score \}\}

Follow this step-by-step reasoning process to write your essay.

STEP\_1: Identify phrases, words, sentences in the Reference essay leading to the Original score.

STEP\_2: Propose minimal edits to those phrases, words, sentences so that the revised essay would receive the Desired score.

STEP\_3: Produce the revised essay by applying the STEP\_2 edits to the Reference essay.

Your goal: rewrite the Reference essay with minimal changes so it would receive the Desired score.

IMPORTANT: The task is NOT complete until the FINAL\_JSON block is printed.

You MUST:

1) Output STEP\_1, STEP\_2, STEP\_3 reasoning.

2) THEN output the FINAL\_JSON block.

3) The FINAL\_JSON block must appear EXACTLY ONCE.

4) The FINAL\_JSON block must be the LAST thing in your response.

5) Do NOT stop after STEP\_3.

6) Do NOT omit the FINAL\_JSON block.

Inside the FINAL\_JSON block:

- Output ONLY valid JSON.

- The JSON must contain exactly one key: ``final\_essay''.

- The value must be a single-line string.

- Do not include newline characters inside the string.

- Do not output any additional text after \texttt{<{}<{}<END\_FINAL\_JSON>{}>{}>}.

SENTINELS (exact):

\texttt{<{}<{}<FINAL\_JSON>{}>{}>}

\{ ``final\_essay'': ``...'' \}

\texttt{<{}<{}<END\_FINAL\_JSON>{}>{}>}
\end{promptbox}
\caption{Example Identify-and-Replace (I\&R) prompt for the DREsS ``Language'' criterion.}
\label{fig:ir_prompt}
\end{figure*}

%% file: prompts/scoring.tex
\begin{figure*}
\begin{promptbox}{Scoring Prompt}
Your task is to give a score for a summary of a passage according to the guidelines given in a rubric.

You will be given a summary that a student wrote for a passage.
For context, you will also be given the passage, which was summarized, and a rubric, which outlines how the summary should be scored for the provided criteria.
You will provide the score and a 1--2 sentence justification in JSON format.
\\ \\
\#\#\# General Guidelines

1.) Make sure to read through the Rubric section below before starting.

2.) You can provide an integer score between 1 and 4 inclusive, that is 1, 2, 3, or 4.

Please make sure you read and understand these instructions carefully. Keep this document open while reviewing, and refer to it as needed.
\\ \\
\#\#\# Output Format

Return your answer as a JSON object with the following structure:

\{

  ``justification'': ``<1--2 sentence justification for your choice>'',

  ``score'': <selected\_score>

\}

- Both fields must be included and non-empty strings. Justification should not exceed two sentences; if multiple lines are needed, concatenate into a single line (no embedded newlines or line breaks).

- Only these two fields should be present in the output JSON.

- Order the JSON fields as shown above: ``justification'' first, then ``label''.

- Both the ``justification'' and ``label'' values must always be strings and must be valid JSON values.

- If the input (summary, passage, or rubric) is missing or empty, output the following JSON: \{ ``justification'': ``ERROR: Cannot score as input information is insufficient or missing.'', ``score'': ``undefined'' \}
\\ \\
\#\#\# Rubric

\#\#\# Criterion: Language Beyond Source Text

\#\#\#\# Description

This criteria measures how well the original aspects of the summary show a mastery of the english language in terms of grammar and vocabulary. The aspects of the summary which are not clearly verbatim copies should be assessed for this criteria. The writers of the summary will be in primary school between grades 3 and 12, so sophisticated, adult-level vocabulary and grammar should not be expected.
\#\#\#\# Possible Scores:

- 1: Poor

- 2: Fair

- 3: Good

- 4: Excellent

\#\#\#\# Score Descriptions:

- **1**: Summary shows a very basic understanding of lexical and syntactic structures.

- **2**: Summary shows an understanding of lexical and syntactic structures.

- **3**: Summary shows an appropriate range of lexical and syntactic structures.

- **4**: Summary shows an excellent range of lexical and syntactic structures.

\#\#\# Passage

\{\{ reading\_passage\_text \}\}

\#\#\# Summary

\{\{ countefactual\_summary \}\}

\#\#\# Score
\end{promptbox}
\caption{Example scoring model prompt for ``Language'' criterion.}
\label{fig:score_prompt}
\end{figure*}